\documentclass[journal]{IEEEtran}
\usepackage{graphicx}
\usepackage{subcaption}
\usepackage{amsmath}
\usepackage{url}
\usepackage{multirow}
\usepackage{orcidlink}
\usepackage[utf8]{inputenc}
\usepackage[T1]{fontenc}
\usepackage{cite}
\usepackage{array}
\usepackage{xcolor}
\usepackage{colortbl}
\usepackage{eso-pic}

\begin{document}
\title{UAV-Based Computer Vision for Cattle Detection and Feed Bunk Spatial Analysis}

\author{Pedro Oliveira Torrente\,\orcidlink{0009-0002-2313-2375},
        Diego Giaretta\,\orcidlink{0009-0003-9279-031X},
        Lucas Pascotti Valem\,\orcidlink{0000-0002-3833-9072},
        Jianglong Yan\,\orcidlink{0000-0002-4693-7072},
        Everton Castel\~ao Tetila\,\orcidlink{0000-0001-5396-766X},
        Jayme Garcia Arnal Barbedo\,\orcidlink{0000-0002-1156-8270},
        and Liang Zhao\,\orcidlink{0000-0002-1502-6604}\thanks{Pedro Oliveira Torrente, Diego Giaretta, Lucas Pascotti Valem, and Jianglong Yan are with the Institute of Mathematics and Computer Science (ICMC), University of São Paulo (USP), São Carlos - SP, Brazil.}\thanks{Everton Castel\~ao Tetila is with the Federal University of Grande Dourados (UFGD), Dourados - MS, Brazil, and also with Embrapa Digital Agriculture, Campinas - SP, Brazil.}\thanks{Jayme Garcia Arnal Barbedo is with Embrapa Digital Agriculture, Campinas - SP, Brazil.}\thanks{Liang Zhao is with the Faculty of Philosophy, Sciences and Letters at Ribeir\~ao Preto (FFCLRP); University of São Paulo (USP), Ribeir\~ao Preto - SP, Brazil.}\thanks{Corresponding author: Lucas Pascotti Valem (e-mail: lucas@icmc.usp.br).}

\thanks{This work involved no animal experimentation, handling, or intervention by the authors. The analysis was conducted exclusively on a pre-existing, publicly available dataset of non-invasive UAV imagery, in which cattle were remotely observed during their normal commercial feedlot routine.}}

\markboth{Preprint of paper accepted}{}

\maketitle
\AddToShipoutPictureFG*{\AtPageUpperLeft{\put(0,-22){\makebox[\paperwidth][c]{\footnotesize\itshape Preprint of paper accepted in IEEE Transactions on AgriFood Electronics}}}}

\begin{abstract}

Beef cattle production is central to Brazilian agribusiness and faces growing demands for efficiency, sustainability, and animal welfare. Precision Livestock Farming (PLF) offers tools to address these challenges, particularly through monitoring feeding-related spatial patterns in confined systems.
Feeding management plays a decisive role in animal performance, yet its evaluation still relies largely on subjective, labor-intensive visual inspections that are hard to scale and prone to bias.
This work proposes a computer vision framework based on Unmanned Aerial Vehicles (UAVs) to quantify cattle alignment and spatial dispersion at the feed bunk under real-world feedlot conditions.
A UAV dataset was collected during flight campaigns at a commercial feedlot in Mato Grosso do Sul, Brazil. Images were manually annotated into cattle head and cattle back classes to support object detection and image-level spatial analysis of cattle distribution. You Only Look Once (YOLO) models, namely YOLOv8 and YOLOv11, were evaluated, with YOLOv8 small used for the downstream spatial analysis. The feed bunk region was segmented using the Segment Anything Model (SAM), and geometric metrics based on detected head positions were developed to characterize alignment, spacing regularity, and clustering.
{The selected detector achieved 84.30\% mAP50 and 35.50\% mAP50-95, with an inference time of approximately 2.7~ms per image on an NVIDIA RTX~5060~Ti GPU.
The resulting metrics distinguished representative regular and irregular spatial distributions} by Root Mean Square Deviation - RMSD (1.06 m and 4.26 m), Coefficient of Variation - CV (42.70\% and 96.31\%), and clustering (0 and 2 clusters), respectively.
These findings indicate that the proposed approach can provide indicators for assessing feed bunk spatial patterns in PLF.

\end{abstract}


\IEEEpeerreviewmaketitle

\section{Introduction}

Beef cattle production is central to Brazilian agribusiness and the global meat supply chain. Brazil ranks among the world's largest beef exporters, accounting for nearly 21\% of global exports and an estimated herd of 194 million animals in 2024, while generating millions of direct and indirect jobs \cite{ani14162340}.
In response to increasing global food demand and the need for sustainable intensification, livestock systems in Brazil have progressively incorporated digital technologies, artificial intelligence, and remote sensing tools to improve production efficiency, environmental sustainability, and animal welfare.

Traditional cattle monitoring remains a labor-intensive activity that demands time, specialized personnel, and substantial operational costs, particularly in extensive and confined production systems. In this context, Precision Livestock Farming (PLF) has emerged as a modern and strategic approach to livestock management. As defined by \cite{10.2527/af.2017.0102}, PLF extends the principles of precision agriculture to animal production by enabling repeated and individualized monitoring~\cite{MHAFF2025}, with the objective of optimizing animal performance, improving resource efficiency, and reducing environmental impacts. Advances in sensing and data acquisition technologies have enabled the collection of large volumes of behavioral and physiological data~\cite{YUAN2026101867}, supporting more accurate and timely management decisions~\cite{RAJAGUKGUK2025101539}.

Among the technologies driving this transition, Unmanned Aerial Vehicles (UAVs) have become increasingly popular due to their ability to rapidly acquire high-resolution images over large areas at relatively low operational expenses \cite{barbedo2018perspectives, barbedo2019review}. When combined with computer vision and Deep Learning (DL), UAV imagery offers a powerful platform for key livestock management tasks, such as automated counting, weight estimation, and monitoring feeding-related spatial indicators \cite{10.1093/af/vfae028, eniac}, while reducing subjectivity and improving the reliability of zootechnical indicators~\cite{XU2024103732, fi17090431}.

In this work, we use a UAV-based dataset collected in a real-world feedlot experiment conducted by the UFGD research group at Campanário Farm, located in Mato Grosso do Sul, Brazil. The imagery was acquired across repeated flight campaigns, providing consistent overhead views. We manually annotated two complementary classes, cattle head and cattle back, to support both effective detection and a more interpretable analysis of animal orientation and spatial distribution at the feed bunk.
In this study, feeding-related spatial indicators refer to image-level, instantaneous descriptors of cattle positions relative to the feed bunk, rather than temporal behavioral measurements such as visit frequency, feeding duration, or individual intake.
Detecting the cattle head as a distinct class is a choice motivated by its relevance to feeding-related spatial interpretation. Head localization provides evidence of animal orientation relative to the feed bunk and proximity to it, while back detections support future applications such as individual identification. This dataset enables evaluation of detection performance and feed bunk alignment metrics under realistic operational constraints.

Feeding management is a key determinant of animal performance in confined systems, and linear feed bunk space plays a central role in shaping competitive interactions among animals.
Insufficient feed bunk space intensifies competition, letting dominant animals restrict access and increasing variability in intake and weight gain \cite{devries2019feeding}, while inadequate management can cause chronic underfeeding (limiting weight gain) or overfeeding (raising waste and ruminal acidosis risk) \cite{pritchard2003bunkManagement}.
Therefore, accurate monitoring of feeding-related spatial distribution along the feed bunk is essential for optimizing nutritional management and ensuring animal welfare.
Recent computer vision approaches have also explored automated monitoring of feed bunks, including individual animal identification and feed bunk condition scoring using overhead cameras and clustering techniques, demonstrating the potential of systems to support feeding management decisions~\cite{skae234.002}.

Within this context, computer vision offers a promising alternative for evaluating cattle alignment and spatial patterns during observed feeding moments~\cite{CHELOTTI2024150}. UAV imagery enables non-invasive, wide-area observation, allowing quantitative assessment of how animals distribute themselves around the feed bunk. This study combines deep learning based object detection with geometric analysis of cattle spatial arrangement. Detectors from the \textit{You Only Look Once} (YOLO) family~\cite{make5040083}, specifically YOLOv8 and YOLOv11~\cite{khanam2024yolov11} in configurations ranging from \textit{nano} to \textit{extra large}, are evaluated for cattle detection in aerial images. A feed bunk segmentation module defines a precise region of interest for more accurate alignment analysis, and geometric dispersion and uniformity metrics are then computed from the detected positions to support a data-driven evaluation of instantaneous feed bunk spatial patterns. The objective is to propose a computer vision approach using UAV imagery to detect cattle and quantify their spatial dispersion and feeding alignment through YOLO-based models and geometric metrics.

The novelty of this work lies in a practical framework that converts standard computer vision outputs into physical, task-specific geometric indicators for instantaneous assessment of cattle distribution along the feed bunk.
The \textbf{main contributions} of this work are summarized as follows:

\begin{itemize}
    \item A UAV-based computer vision framework for non-invasive and image-level assessment of cattle spatial distribution at the feed bunk in UAV imagery.
    \item A comparative evaluation of multiple YOLO-based object detectors (YOLOv8 and YOLOv11, across model sizes) applied to cattle detection under real farm conditions.
    \item A feed bunk segmentation module that defines a precise region of interest for analyzing cattle alignment and spatial distribution at the feed bunk.
    \item The development of geometric dispersion and uniformity metrics to quantify cattle distribution along the feed bunk, supporting data-driven monitoring of feed bunk occupation.
    \item Experimental evidence that the proposed metrics quantify and discriminate contrasting spatial occupation patterns at the feed bunk, providing a basis for future decision-support tools in feeding management.
    \end{itemize}

The remainder of this paper is organized as follows.
Section~\ref{sec:relatedWork} reviews related work on UAV-based livestock monitoring, YOLO-based cattle detection, segmentation models, and feed bunk analysis. Section~\ref{sec:methods} describes the proposed methodology, including the dataset preparation, the training and comparison of YOLO detectors, the feed bunk segmentation procedure, and the geometric metrics used to quantify cattle alignment and dispersion. Section~\ref{sec:experimental-evaluation} reports the experimental results, covering both effectiveness and inference time analyses and the evaluation of instantaneous feed bunk spatial patterns using the proposed metrics. Finally, Section~\ref{sec:conclusion} concludes the paper and outlines directions for future work.

\section{Related Work}
\label{sec:relatedWork}

Precision Livestock Farming has increasingly used computer vision, remote sensing, and machine learning to reduce the dependence on subjective visual inspection in livestock management~\cite{norton2019digitalRepresentations,bretas2024grazinglandPLF}. In feedlot systems, this is particularly relevant because feeding space, bunk design, and competitive interactions influence feeding behavior, intake variation, and animal welfare~\cite{pritchard2003bunkManagement,devries2019feeding}. UAV imagery has emerged as a scalable and non-invasive alternative for livestock monitoring, enabling cattle detection, counting, trough reading, production monitoring, and growth assessment under practical farm conditions~\cite{barbedo2020counting,bastos2024drones,wang2025uavlidar,yan2026morphology}. However, most UAV-based studies focus on detection, counting, or morphometric analysis~\cite{bhujel2025}, while the geometric relationship between cattle positions and the feed bunk during feeding events remains less explored.

YOLO-based detectors have been widely used in agricultural and livestock applications due to their balance between detection accuracy and inference speed~\cite{badgujar2024yoloAgriculture}, including cattle body detection and feeding-activity recognition~\cite{qiao2023yolov5asff,guarnidolopez2024yolo}. Segmentation foundation models such as the Segment Anything Model (SAM) have also been explored in agricultural and remote-sensing scenarios~\cite{Kirillov_2023_ICCV,li2023samadapter,gui2024samRemoteSensing}. Unlike studies that use detection or segmentation as final outputs, this work uses cattle-back and cattle-head detections and feed bunk segmentation as geometric references for downstream spatial analysis.

{Cattle detection has also been combined with tracking to monitor individual animals over time. Mar et al.~\cite{mar2023cow} combined animal location and visual appearance to maintain individual identities across video frames, including after temporary occlusions. Moe et al.~\cite{moe2025minimizing} developed a system for tracking black cattle during the day and at night, reducing model size and processing time. These studies focus on individual cattle tracking in videos from fixed cameras, while our framework quantifies instantaneous cattle distribution relative to the feed bunk in UAV images.}

Automated feed bunk monitoring is the closest line of work to this study. Previous approaches have addressed computer vision-based feed bunk management, bunk-score datasets, and automated bunk-score classification~\cite{dorea2019feedBunkCV,depaula2023feedBunkDataset,dasilva2025feedBunkScore}. Related works by Pi et al.~\cite{pi2024feedGrading} and Mendes et al.~\cite{skae234.002} further combine computer vision, individual animal identification, feed grading, and clustering analysis to support feeding management.
The present study differs by using UAV imagery instead of fixed or close-range cameras, focusing on group-level instantaneous spatial distribution rather than individual intake, visit detection, or feed-leftover classification, and integrating cattle-back and cattle-head detection, feed bunk localization using SAM, and geometric metrics to quantify transversal proximity, longitudinal spacing uniformity, and clustering. Therefore, the proposed framework complements existing feed bunk scoring and animal identification methods with interpretable spatial indicators of cattle distribution during feeding.

\section{Methodology}
\label{sec:methods}

This section describes the proposed method for analyzing cattle spatial distribution relative to the feed bunk using aerial imagery and deep learning. The method integrates object detection, semantic segmentation, and geometric analysis to quantify cattle alignment, proximity, and longitudinal spacing along the feed bunk.

\subsection{Overview}

\begin{figure}[!htbp]
 \centering
 \includegraphics[width=.49\textwidth]{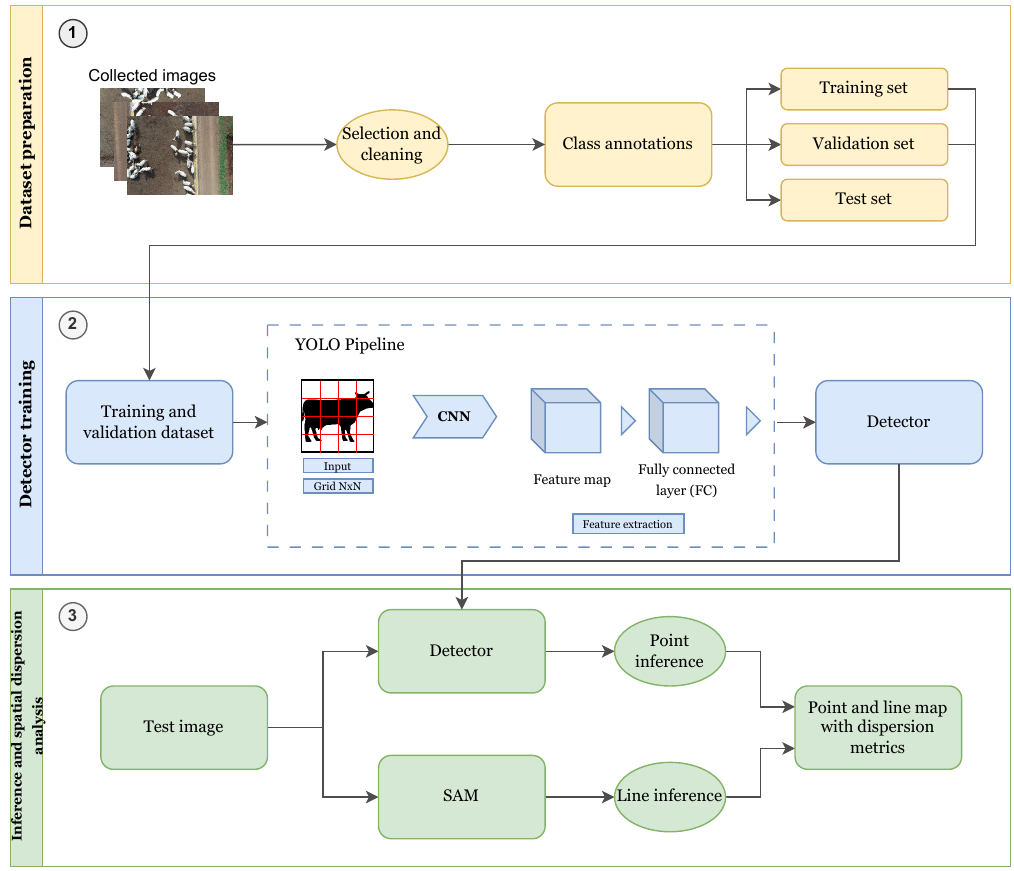}
  \centering
 \caption{Flowchart of the proposed methodology.}
 \label{figura:fluxograma}
\end{figure}

The proposed workflow transforms UAV images into a geometric representation of cattle distribution at the feed bunk. Detected head positions and a segmented feed bunk reference are combined to compute distance and spacing-based metrics that quantify alignment, dispersion, and clustering during observed feeding moments.
Figure~\ref{figura:fluxograma} summarizes the proposed method, which is structured into three main stages:
\begin{enumerate}
    \item \textbf{Dataset preparation:} selection, cleaning, and dual-class annotation of UAV imagery (\textit{cattle-head} and \textit{cattle-back}) to improve robustness to posture and orientation variations;
    \item \textbf{Detector training:} training and comparative evaluation of multiple YOLO variants~\cite{make5040083}, followed by the adoption of YOLOv8 \textit{small} as the detector for the downstream spatial analysis due to its competitive effectiveness and stable implementation.
    \item \textbf{Inference and spatial analysis:} cattle localization, feed bunk segmentation using the \textit{Segment Anything Model} (SAM)~\cite{Kirillov_2023_ICCV}, and quantification of spatial dispersion through geometric and statistical metrics.
\end{enumerate}

A detailed description and discussion of each stage are provided in the corresponding subsections.

\subsection{Dataset Preparation}
\label{sec:preparacao_dataset}

\subsubsection{Dataset}
\label{sec:conjunto_dados}

The dataset used in this study was obtained through a field experiment conducted at Campanário Farm, located in the municipality of Caarapó, Mato Grosso do Sul, Brazil. Data acquisition was carried out by a team coordinated by Prof. Dr. Everton Castelão Tetila (UFGD)~\footnote{The dataset is publicly available and can be downloaded at \\\url{https://github.com/EvertonTetila/NelloreBeefCattleDataset}.}. This dataset has previously been used in studies on cattle morphology and growth analysis~\cite{yan2026morphology} and mounting behavior detection~\cite{11578135}. In this work, the dataset is used to support feed bunk spatial analysis through cattle-head and cattle-back detection and geometric alignment metrics.

The experiment involved repeated monitoring of a group of 110 Nelore cattle (\textit{Bos indicus}), kept in a specific corral at the farm from July 10 to October 27, 2024. Each animal was individually identified by a numeric hot-iron brand on the dorsal region, enabling individual traceability throughout all image acquisition campaigns. The farm hosts a total herd of approximately 14,600 cattle of different breeds, which introduces additional challenges for detection methods due to morphological and chromatic variability among animals, directly affecting visual contrast and detection consistency.

Image acquisition was performed using a multirotor unmanned aerial vehicle (UAV), specifically a \textit{DJI Phantom 4 Advanced} equipped with a 20~MP, 1-inch Sony sensor. Images were captured at a native resolution of 4864~$\times$~3648 pixels (4:3, $\approx$17.7~MP).

The flight campaigns were conducted periodically, typically every one to three weeks. In each campaign, images were captured at an altitude of 15 meters above ground level. This altitude was selected because it provides a broader view of the confinement area, favoring spatial analysis of cattle dispersion relative to the feed bunk.

{The Ground Sampling Distance (GSD) was estimated using image metadata extracted with ExifTool and the camera specifications. Given the camera's pixel pitch of approximately $p = 2.41~\mu$m, a real focal length of $f = 8.8$~mm, and a recorded flight altitude of approximately $H = 15$~m, the resulting GSD is approximately $4$~mm/px (i.e., $1$~px $\approx 0.4$~cm at ground level), computed as $\text{GSD} = p\,H/f$. Because the recorded altitude} is referenced to the UAV takeoff point and the terrain is not perfectly level, the height above ground, and therefore the GSD, may vary slightly across the scene and between campaigns. Therefore, we adopt $\approx 4$~mm/px as an approximate conversion factor used to express the geometric metrics in physical units.

\subsubsection{Image Selection and Cleaning}

Each captured image contains multiple cattle, representing different grouping formations and positioning patterns relative to the feed bunk. The selected images were divided into three subsets according to the chronological acquisition order: training (July 10, July 31, and August 11), validation (August 21, August 31, September 21, and September 28), and test (October 5, October 12, October 19, and October 27). This chronological split strategy was adopted to reduce temporal and campaign-related bias, while ensuring that images from the same acquisition campaign were not shared across subsets. Therefore, the test set contains later acquisition conditions that were not observed during training or validation.

During this phase, image cleaning was also performed by removing records of farm regions without cattle and discarding images captured at the beginning of flights, where significant variations in UAV altitude occurred.

\subsubsection{Class Annotation}

After image selection, the open-source software \textit{LabelMe} was used to annotate regions of interest corresponding to two classes: \textit{cattle-back} and \textit{cattle-head}. This dual-class strategy represents an alternative to single-class cattle detection. Using a single cattle class poses challenges due to variations in posture, orientation, and morphology, which can introduce ambiguities in detections and hinder identification. Therefore, the adoption of two classes is a core aspect of the proposed methodology, enabling discrimination between the dorsal region and the head of cattle, supporting more detailed analyses of orientation and positioning, and expanding the potential applicability of the model in future studies. In particular, separating cattle-head and cattle-back annotations improves the interpretability of the detections for feed bunk spatial analysis: head detections indicate proximity to the feed bunk and, when paired with back detections, may enable orientation analysis in future work.

In commercial livestock systems, such as Campanário Farm, the dorsal region of animals is commonly marked with a unique numeric identifier for individual tracking until slaughter. Thus, detecting only the dorsal region enables automatic cropping of these areas for future applications in animal identification using alternative computer vision techniques. Detecting the head, in turn, is essential for analyzing cattle distribution relative to the feed bunk, as it indicates the animal’s orientation. For instance, a bovine facing away from the feed bunk may indicate irregular feed distribution at that location. Additionally, this approach reduces background content within regions of interest and decreases class variability, favoring model performance and convergence.

The distribution of annotations across each dataset subset is presented in Table~\ref{tabela:classes}. 
The resulting dataset comprises 232 images and 5,027 manually labeled instances, divided into 97 training images, 47 validation images, and 88 test images. 
The expanded dataset is publicly available, and the official
repository has been updated with Version 2, including the revised annotations and
the chronological train/validation/test split used in this study.

\begin{table}
\centering
\caption{Distribution of annotated instances across the training, validation, and test subsets, considering cattle-head (head region indicating orientation relative to the feed bunk) and cattle-back (dorsal region). Values are the number of annotated instances (i.e., individual bounding-box labels).}
\label{tabela:classes}
\resizebox{.4\textwidth}{!}{
\begin{tabular}{l|c|c|c}
\hline
\multicolumn{1}{c|}{\textbf{Subset}} & \textbf{Both} & \textbf{cattle-back} & \textbf{cattle-head} \\ \hline
Training   & 1,911 & 988  & 923 \\ \hline
Validation & 1,137 & 579  & 558 \\ \hline
Test       & 1,979 & 1,007 & 972 \\ \hline
\textbf{Total} & \textbf{5,027} & \textbf{2,574} & \textbf{2,453} \\ \hline
\end{tabular}
}
\end{table}

Although the total number of images is moderate, the dataset includes many labeled cattle instances from different UAV acquisition campaigns under real commercial feedlot conditions. These images capture natural variations in animal spacing, occlusion, illumination, and background appearance.

\subsection{Detector Training}
\label{sec:treinamento_detectores}

This stage comprises the training and comparison of two YOLO-family detectors widely used in industry and academia: YOLOv8, an established and commonly adopted version, and YOLOv11. All available model size variants (\textit{nano}, \textit{small}, \textit{medium}, \textit{large}, and \textit{extra-large}) were considered to evaluate the trade-off between effectiveness and computational efficiency.

Training hyperparameters were defined to standardize comparison conditions while balancing performance and accessible computational power. Preliminary tests were conducted using YOLOv8 \textit{nano} to fine-tune parameters.
Due to the high resolution of the images and the relatively small dataset size, an input image size of 640$\times$640 pixels was adopted. This resolution preserves sufficient visual detail for detecting small objects, such as the \textit{cattle-head} and \textit{cattle-back} classes, while remaining feasible in low-compute environments. The batch size was set to 16, ensuring good convergence of loss curves without exceeding processing capacity. A maximum of 50 training epochs was defined, which proved sufficient for the dataset size, avoiding excessive iterations and reducing the risk of overfitting. Additionally, a safety patience of 15 epochs was established to interrupt training if no significant performance improvement was observed.

To perform the training and inference-time evaluation, we used a dedicated workstation equipped with an AMD Ryzen~7~9700X CPU, 32~GB of DDR5-5600 memory, and an NVIDIA GeForce RTX~5060~Ti GPU with 16~GB of GDDR7 memory. 

\subsection{Inferences and Spatial Dispersion Analysis}
\label{sec:inferencias_dispersao}

\begin{figure*}[h!]
    \centering
    \includegraphics[
        width=.84\linewidth,
        trim={0 0 0 2mm},
        clip
    ]{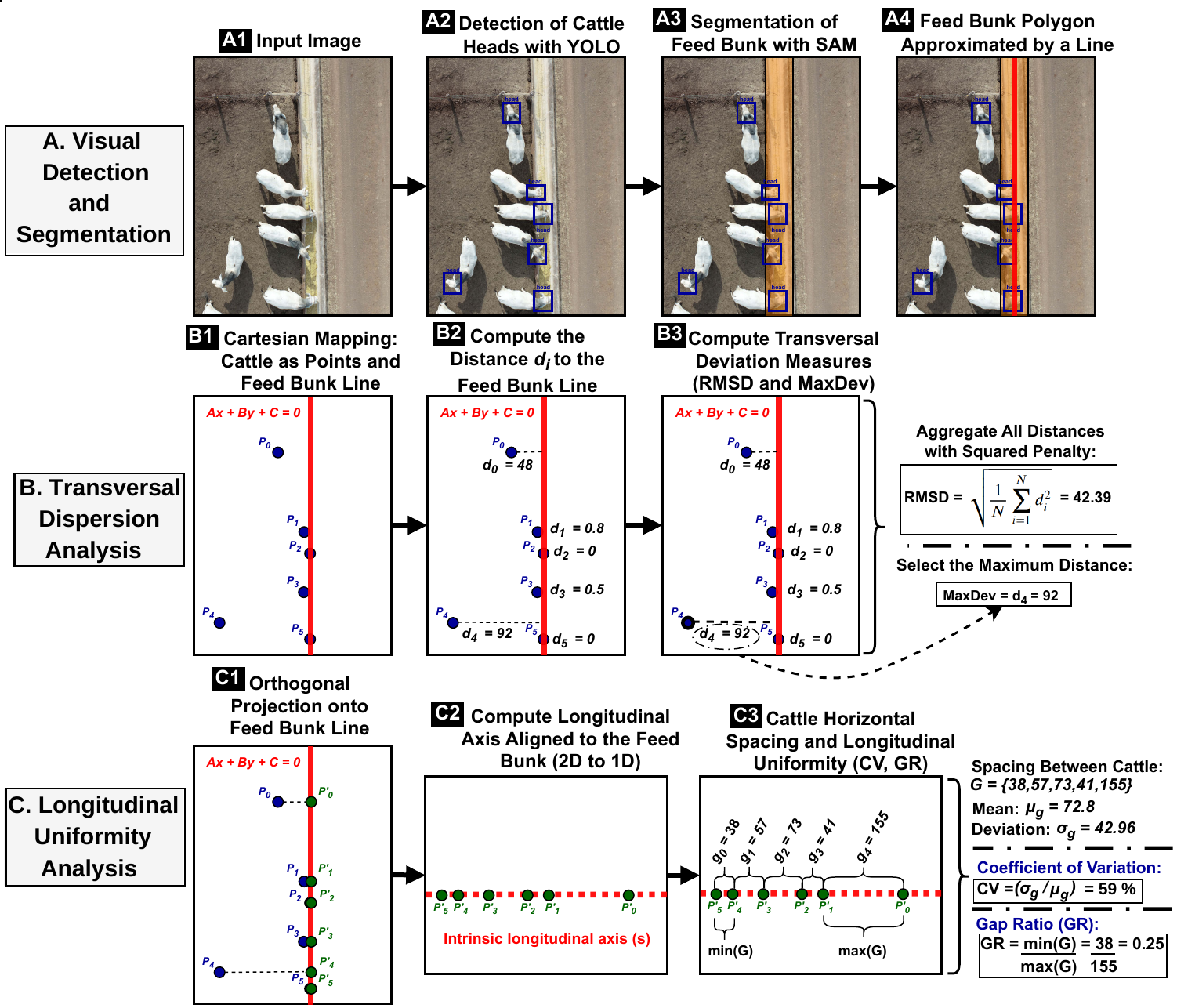}
    \caption{Overview of the steps for the inference and spatial dispersion analysis. Cattle heads are detected using YOLO, the feed bunk is segmented using SAM, the segmentation polygon is approximated by a straight line, and geometric measures are computed based on the Cartesian mapping of bounding box centers.}
    \label{fig:feedbunk_analysis}
\end{figure*}

The objective of this stage is to analyze the dispersion and alignment of cattle relative to the feed bunk using geometric and statistical measures, such as the distance from a point to a line and the distance between the projections of points onto that line. Images are represented as a Cartesian plane, where one unit corresponds to one image pixel.
Figure~\ref{fig:feedbunk_analysis} presents an overview of the proposed inference and spatial dispersion analysis workflow.

For the downstream spatial analysis, YOLOv8 \textit{small} was adopted as the detection model. Although other variants achieved slightly higher detection metrics, YOLOv8 \textit{small} provided a favorable balance between detection effectiveness and implementation stability. Its consolidated implementation makes it suitable for the proposed analysis pipeline.

Using this detector, the analyzed UAV images were inferred, with detection and labeling of bounding boxes corresponding to each cattle class. During inference, non-maximum suppression (NMS) with an IoU threshold of 0.10 was applied to remove overlapping detections belonging to the same class before generating the final bounding boxes used in the spatial analysis. The center point of each bounding box was mapped onto the Cartesian plane to support statistical analyses of cattle location relative to the feed bunk.

\subsubsection{Feed Bunk Segmentation and Line Equation}

Initially, it was necessary to identify the spatial position of the feed bunk in each image for subsequent geometric analysis. For this purpose, the Segment Anything Model (SAM) was employed as a segmentation tool due to its ability to efficiently isolate objects with minimal manual intervention. In this work, the \texttt{vit\_h} version of SAM was used.

An approximate indication of the feed bunk region was provided to the model, allowing SAM to generate a binary mask that approximately delineates the feed bunk region, from which the reference line is derived in the image. This mask was then used to extract reference points required to compute the line equation representing the feed bunk orientation in the image’s Cartesian plane.

The vertical central segment of this polygon was approximated by a straight line in the general form:
$A x + B y + C = 0$, where \(A\), \(B\), and \(C\) are coefficients obtained from a linear fit of the vertical midpoint coordinates of the polygon.
{This formulation assumes a straight feed bunk, as observed in the evaluated images, while other layouts (e.g., curved or L-shaped) would require adapting the geometric reference and the corresponding distance and projection calculations.}

\subsubsection{Transversal Dispersion Measures (Proximity)}

To analyze cattle offset relative to the feed bunk, the perpendicular distance ($d_i$) between each inferred cattle point (i.e., cattle heads) and the line representing the feed bunk is used, a fundamental concept in analytical geometry~\cite{lay2016linear}. The distance $d_i$ is defined as:

\begin{equation}
\label{eq:distancia-ponto-reta}
    d_i = \frac{|A x_i + B y_i + C|}{\sqrt{A^2 + B^2}}.
\end{equation}

To quantify dispersion, mean and standard deviation alone are insufficient, as measures that penalize large deviations, typical of misalignment or \textit{outliers}, are required. The aggregated measures adopted are:

\begin{itemize}
    \item \textbf{Root Mean Square Deviation (RMSD):} A standard metric in statistics and regression analysis that measures average error magnitude by squaring individual distances, disproportionately penalizing points far from the line.

    \begin{equation}
    \label{eq:rmsd}
        \text{RMSD} = \sqrt{\frac{1}{N}\sum_{i=1}^N d_i^2}
    \end{equation}

    \item \textbf{Maximum Absolute Deviation ($\text{Max Dev}$):} Defined as $\text{Max Dev} = \max(d_i)$, this metric captures the worst-case deviation, ensuring tolerance limits are respected.
\end{itemize}

\subsubsection{Longitudinal Uniformity Measures (Spacing)}

To analyze longitudinal distribution along the feed bunk, the 2D problem is reduced to a 1D formulation by defining an intrinsic coordinate $s$ along the line. This is achieved by orthogonally projecting inferred cattle points (i.e., cattle heads) onto the computed line. Based on linear algebra~\cite{lay2016linear}, each point \(P_i(x_i, y_i)\) is mapped to \(P_i'(x_i', y_i')\) on the line as:

\begin{equation}
\label{eq:proj-ortogonal}
\left\{
\begin{aligned}
    x_i' &= \dfrac{B (B x_i - A y_i) - A C}{A^2 + B^2}, \\[8pt]
    y_i' &= \dfrac{A (-B x_i + A y_i) - B C}{A^2 + B^2}.
\end{aligned}
\right.
\end{equation}

A direction vector \(\vec v = (-B, A)\), perpendicular to the normal vector \(\vec n = (A,B)\), defines the orientation along the feed bunk line. Its normalized form gives the unit direction vector:
\begin{equation}
\label{eq:unit-direction}
    \vec u_v = \frac{\vec v}{\lVert \vec v \rVert} = \frac{(-B,\, A)}{\sqrt{A^2 + B^2}}.
\end{equation}

Selecting a reference point $P_{ref}$ (e.g., the first projected point), the longitudinal coordinate $s_i$ is obtained by projecting $\vec{P_{ref}P'_i}$ onto the unit direction vector $\vec{u}_v$:

\begin{equation}
\label{eq:conjunto-s}
    s_i = \vec{P_{ref}P'_i} \cdot \vec{u}_v.
\end{equation}

The ordered set $S = \{s_1, s_2, \ldots, s_N\}$ enables computation of physical spacings (\textit{gaps}) between neighboring animals: {$g_j = s_{(j+1)} - s_{(j)}$, for $j=1,\ldots,N-1$, with $G = \{g_1, g_2, \ldots, g_{N-1}\}$ denoting the set of gaps.}

The following relative metrics are used to assess longitudinal uniformity:

\begin{itemize}
    \item \textbf{Coefficient of Variation ($\text{CV}$):} Defined as the ratio between the standard deviation ($\sigma_g$) and mean ($\mu_g$) of the spacings. Ideal uniformity corresponds to $\text{CV} \to 0\%$.

    \begin{equation}
        \text{CV} = \frac{\sigma_g}{\mu_g} \times 100\%.
    \end{equation}

    \item \textbf{Gap Ratio ($\text{GR}$):} A geometric metric defined as the ratio between the smallest and largest spacing:
    \begin{equation}
        \text{GR} = \frac{\min(G)}{\max(G)}.
    \end{equation}
    Values approaching 1 indicate similar gap sizes, while values near 0 indicate strong spacing imbalance.
\end{itemize}

\subsubsection{Qualitative Clustering Analysis}

To qualitatively validate clustering along the feed bunk, the \textbf{DBSCAN} (\textit{Density-Based Spatial Clustering of Applications with Noise}) algorithm is applied to the longitudinal coordinates $s_i$. DBSCAN identifies clusters by detecting regions of high point density~\cite{Ester1996}. It is suitable for this context because it does not require a predefined number of clusters and can handle noise.

In this analysis, the neighborhood radius $\epsilon$ is defined as $50\%$ of the mean inter-animal spacing, and a minimum of three points is required to form a cluster. Points that do not satisfy these criteria are classified as noise.
From this procedure, a categorical metric is obtained corresponding to the number of clusters identified along the feed bunk, excluding noise points.
Ideally, a uniform animal distribution results in {no clusters}, whereas the occurrence of {one or more clusters} indicates aggregation and non-uniform longitudinal spacing.

This framework decouples the analyses by allowing the error $d$ to quantify transversal dispersion (animal proximity to the feed bunk), while the spatial distribution of the $s_i$ coordinates characterizes longitudinal uniformity. Together, these complementary measures enable a quantitative characterization of animal distribution and may support future correlation analyses with feed uniformity.

\section{Experimental Evaluation}
\label{sec:experimental-evaluation}

This section presents the experimental evaluation of the proposed UAV-based framework. First, a comparative analysis of YOLOv8 and YOLOv11 detectors is conducted in Section~\ref{sec:comparative-detectors}, including an effectiveness analysis based on mean Average Precision (mAP) metrics (Section~\ref{sec:acc-analysis}) and an inference time comparison (Section~\ref{sec:time_inf}). Next, the inference stage is detailed in Section~\ref{sec:inferencia}, covering model selection. Finally, Section~\ref{sec:feedbunk} discusses feed bunk mask extraction using SAM, cattle localization, and the evaluation of the proposed geometric metrics for quantifying transversal dispersion, longitudinal spacing, and clustering patterns.

\subsection{Detector Comparison}
\label{sec:comparative-detectors}

To perform the comparison between the two YOLO detector families, the post-training performance of each version was analyzed, considering the mAP50 and mAP50-95 metrics relative to the total number of parameters of each model. The behavior of these same metrics with respect to inference speed was also evaluated. Based on this comparison, the detector version that best balances effectiveness and efficiency is selected, in accordance with the scope of this study, and is subsequently used in the inference and cattle dispersion analysis presented in Section~\ref{sec:inferencia}.

\subsection{Effectiveness Analysis}
\label{sec:acc-analysis}

The effectiveness of the trained models can be assessed through the mAP50 and mAP50-95 metrics, which represent the mean average precision computed at an IoU threshold of 50\% and averaged over IoU thresholds from 50\% to 95\%, respectively. In detectors with a larger number of neural network parameters, there is a general tendency for these metrics to increase. This behavior can be observed in the obtained curves.

\begin{figure*}[!ht]
 \centering
 \includegraphics[width=.99\textwidth, clip]{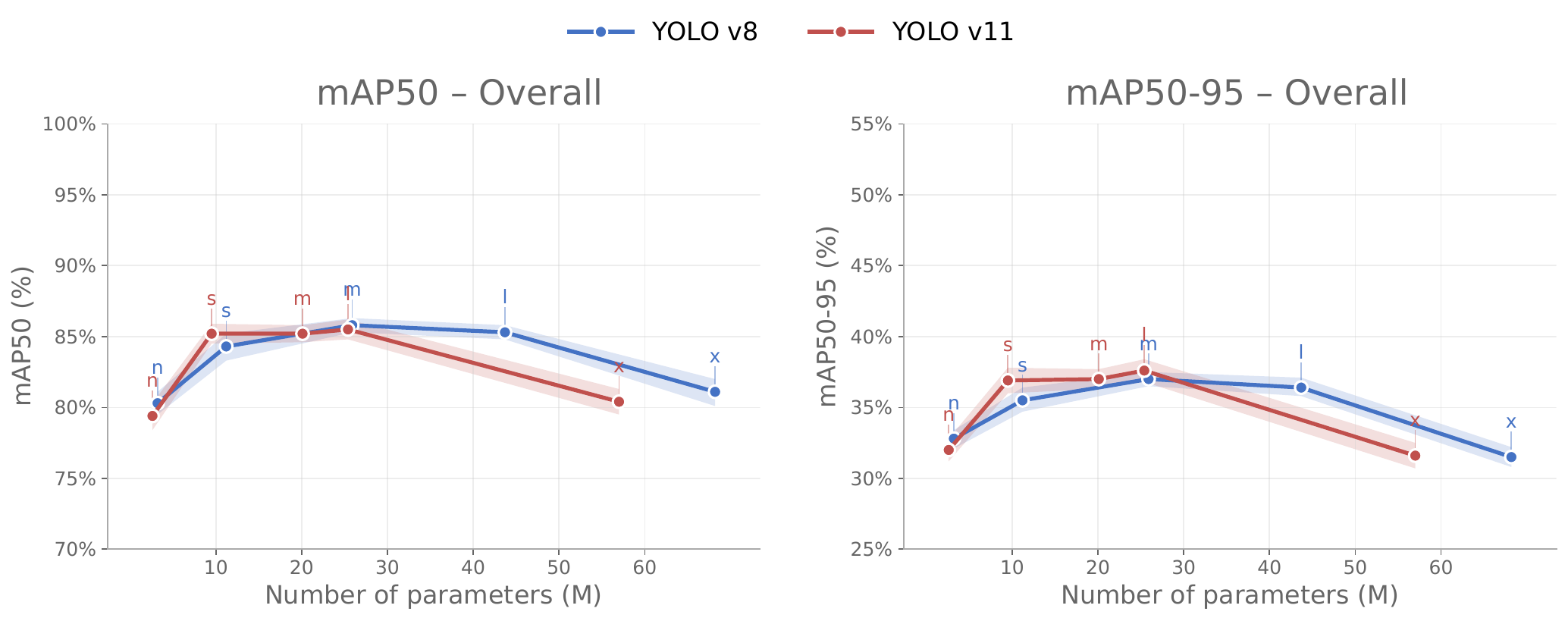}
 \caption{Effectiveness curves as a function of the number of parameters for each model.}
 \label{figura:curvas-map}
\end{figure*}

The mAP50 and mAP50-95 values for each object class and the overall average are presented in Tables~\ref{tabela:map50} and~\ref{tabela:map95}, respectively.
The metric values are reported as the mean and 95\% confidence interval over 10 training runs.
{{Figure~\ref{figura:curvas-map} shows the overall detection performance as a function of the number of parameters for each YOLO variant.}}

\begin{table}[!htbp]
\centering
\caption{Average mAP50 for different YOLO variants.}
\label{tabela:map50}
\resizebox{\columnwidth}{!}
{
\begin{tabular}{l|l|c|c|c}
\hline
\textbf{Model} & \textbf{Param.} & \textbf{Overall} & \textbf{\textit{cattle-back}} & \textbf{\textit{cattle-head}} \\
\textbf{} & \textbf{~~(M)} & \textbf{(\%)} & \textbf{Class (\%)} & \textbf{Class (\%)} \\ \hline
\hline
\textbf{v8n}    & 3.2                 & 80.30 $\pm$ 0.80        & 89.80 $\pm$ 0.50           & 70.70 $\pm$ 1.60           \\ \hline
\textbf{v8s}    & 11.2                & 84.30 $\pm$ 1.00        & 91.70 $\pm$ 0.40           & 76.80 $\pm$ 2.10           \\ \hline
\textbf{v8m}    & 25.9                & 85.80 $\pm$ 0.50        & 91.40 $\pm$ 0.50           & 80.20 $\pm$ 0.80           \\ \hline
\textbf{v8l}    & 43.7                & 85.30 $\pm$ 0.50        & 90.70 $\pm$ 0.60           & 79.80 $\pm$ 0.80           \\ \hline
\textbf{v8x}    & 68.2                & 81.10 $\pm$ 1.00        & 90.00 $\pm$ 0.80           & 72.20 $\pm$ 1.70           \\ \hline\hline
\textbf{v11n}   & 2.6                 & 79.40 $\pm$ 1.00        & 89.10 $\pm$ 0.40           & 69.60 $\pm$ 1.90           \\ \hline
\textbf{v11s}   & 9.5                 & 85.20 $\pm$ 0.70        & 91.50 $\pm$ 0.50           & 78.90 $\pm$ 1.20           \\ \hline
\textbf{v11m}   & 20.1                & 85.20 $\pm$ 0.60        & 91.60 $\pm$ 0.40           & 78.80 $\pm$ 1.10           \\ \hline
\textbf{v11l}   & 25.4                & 85.50 $\pm$ 0.70        & 91.30 $\pm$ 0.60           & 79.60 $\pm$ 1.20           \\ \hline
\textbf{v11x}   & 57.0                & 80.40 $\pm$ 0.90        & 90.30 $\pm$ 0.50           & 70.50 $\pm$ 1.60           \\ \hline
\end{tabular}
}
\end{table}

\begin{table}[!htbp]
\centering
\caption{Average mAP50-95 for different YOLO variants.}
\label{tabela:map95}
\resizebox{\columnwidth}{!}{\begin{tabular}{l|l|c|c|c}
\hline
\textbf{Model} & \textbf{Param.} & \textbf{Overall} & \textbf{\textit{cattle-back}} & \textbf{\textit{cattle-head}} \\
\textbf{} & \textbf{~~(M)} & \textbf{(\%)} & \textbf{Class (\%)} & \textbf{Class (\%)} \\ \hline
\hline
\textbf{v8n}    & 3.2                 & 32.80 $\pm$ 0.60        & 41.30 $\pm$ 0.70           & 24.20 $\pm$ 0.80           \\ \hline
\textbf{v8s}    & 11.2                & 35.50 $\pm$ 0.90        & 42.80 $\pm$ 0.70           & 28.20 $\pm$ 1.30           \\ \hline
\textbf{v8m}    & 25.9                & 37.00 $\pm$ 0.50        & 43.30 $\pm$ 0.50           & 30.70 $\pm$ 0.70           \\ \hline
\textbf{v8l}    & 43.7                & 36.40 $\pm$ 0.70        & 42.20 $\pm$ 0.80           & 30.60 $\pm$ 0.80           \\ \hline
\textbf{v8x}    & 68.2                & 31.50 $\pm$ 0.70        & 37.70 $\pm$ 0.70           & 25.30 $\pm$ 1.10           \\ \hline\hline
\textbf{v11n}   & 2.6                 & 32.00 $\pm$ 0.80        & 40.20 $\pm$ 0.60           & 23.90 $\pm$ 1.10           \\ \hline
\textbf{v11s}   & 9.5                 & 36.90 $\pm$ 0.90        & 43.80 $\pm$ 0.80           & 30.10 $\pm$ 1.00           \\ \hline
\textbf{v11m}   & 20.1                & 37.00 $\pm$ 0.70        & 43.70 $\pm$ 0.90           & 30.30 $\pm$ 0.90           \\ \hline
\textbf{v11l}   & 25.4                & 37.60 $\pm$ 0.80        & 44.00 $\pm$ 0.70           & 31.20 $\pm$ 1.00           \\ \hline
\textbf{v11x}   & 57.0                & 31.60 $\pm$ 0.90        & 38.40 $\pm$ 0.90           & 24.80 $\pm$ 1.00           \\ \hline
\end{tabular}
}
\end{table}

\begin{figure}[!ht]
 \centering
 \includegraphics[width=.49\textwidth]{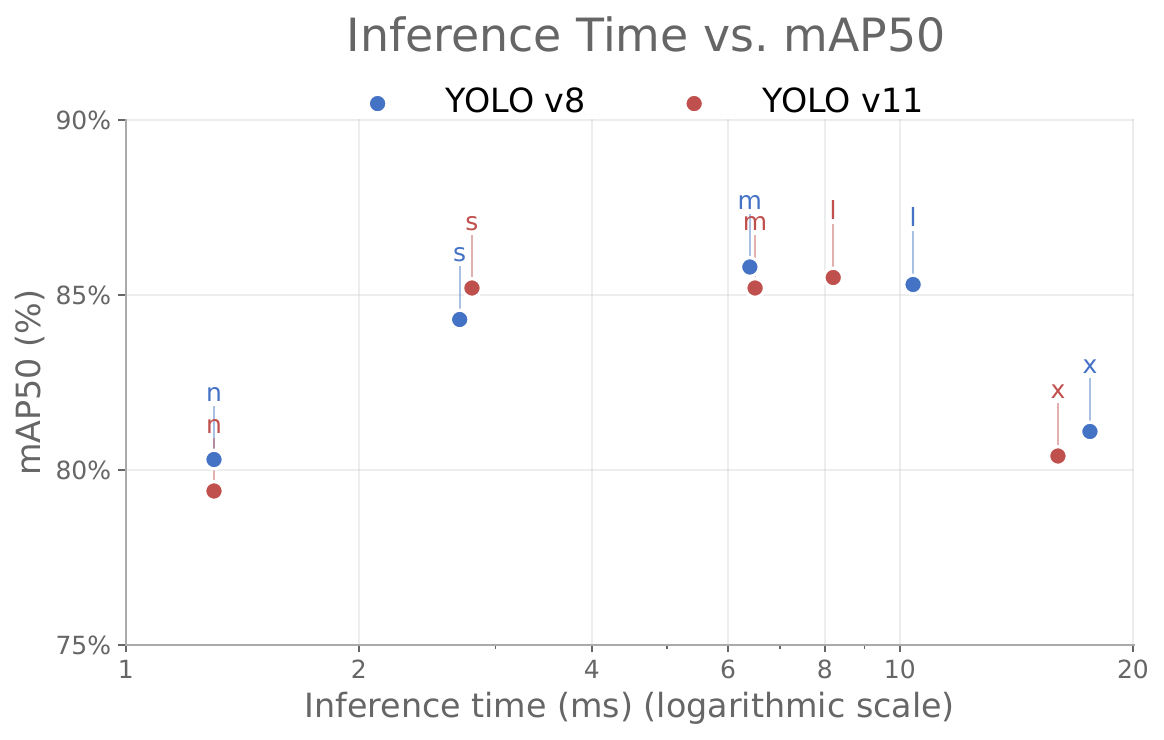}
  \caption{Comparison between effectiveness and inference time.}
 \label{figura:velocidade}
\end{figure}

It is observed that mAP50 increases sharply in the initial model versions, particularly during the transition from the \textit{nano} to the \textit{small} variants. This significant increase likely occurs because, at this stage, the addition of parameters effectively expands the model’s representational capacity, allowing it to learn more complex patterns present in the data and consequently improve performance.

In the \textit{medium} and \textit{large} variants, the growth of mAP50 becomes less pronounced, suggesting that the models approach a saturation point where added complexity yields limited gains. The \textit{extra-large} variants show a decline rather than a plateau, with v8x and v11x falling below the \textit{small} models in both mAP50 and mAP50-95. This is consistent with the relatively small dataset, in which higher-capacity models are more prone to overfitting and thus generalize less well to the chronologically distinct test campaigns.

To assess the impact of the dual-class formulation, an ablation study was conducted with the \textit{small} models (YOLOv8s and YOLOv11s), comparing the proposed two-class training with single-class models trained separately for \textit{cattle-head} and \textit{cattle-back}. As shown in Table~\ref{table:ablation}, the two-class formulation achieved comparable or slightly superior effectiveness in most cases, including under mAP50-95, with differences generally within the confidence intervals. These results indicate that jointly training both classes does not degrade detection performance. The main motivation for the dual-class formulation is therefore interpretability: distinguishing head and back regions provides spatial cues about animal positioning relative to the feed bunk and supports future orientation analysis and individual identification (as described in Section~\ref{sec:preparacao_dataset}).

\begin{table}[!htbp]
\centering
\caption{Ablation study: two-class vs.\ single-class training for \textit{cattle-back} and \textit{cattle-head} detection.}
\label{table:ablation}
\resizebox{\columnwidth}{!}{\begin{tabular}{l|c|c|c|c|c}
\hline
\textbf{Model} & \textbf{Setting} & \multicolumn{2}{c|}{\textbf{mAP50 (\%)}} & \multicolumn{2}{c}{\textbf{mAP50-95 (\%)}} \\ \cline{3-6}
\textbf{} & \textbf{} & \textit{cattle-back} & \textit{cattle-head} & \textit{cattle-back} & \textit{cattle-head} \\ \hline \hline
\multirow{2}{*}{\textbf{v8s}}  & Two-class    & \textbf{91.70 $\pm$ 0.40} & 76.80 $\pm$ 2.10 & \textbf{42.80 $\pm$ 0.70} & \textbf{28.20 $\pm$ 1.30} \\
                               & Single-class & 91.10 $\pm$ 0.40 & \textbf{77.20 $\pm$ 0.50} & 42.30 $\pm$ 0.50 & 27.40 $\pm$ 0.50 \\ \hline
\multirow{2}{*}{\textbf{v11s}} & Two-class    & \textbf{91.50 $\pm$ 0.50} & \textbf{78.90 $\pm$ 1.20} & \textbf{43.80 $\pm$ 0.80} & \textbf{30.10 $\pm$ 1.00} \\
                               & Single-class & 91.40 $\pm$ 0.20 & 77.90 $\pm$ 1.00 & 43.60 $\pm$ 0.60 & 28.80 $\pm$ 0.90 \\ \hline
\end{tabular}
}
\end{table}

\subsection{Inference Time Comparison}
\label{sec:time_inf}

Another important metric is the time required to perform inference on each image. After training, the resulting model can be deployed in real-world applications, often on embedded devices that perform inference locally.

Given the importance of this metric for model selection, an analysis was conducted comparing model effectiveness (in terms of mAP50) with inference time. The corresponding graph is shown in Figure~\ref{figura:velocidade}.

Although embedded deployment was not experimentally evaluated in this study, previous work has shown that lightweight YOLO models can satisfy real-time constraints on NVIDIA Jetson edge devices~\cite{electronics14030638}.

\subsection{Inference Stage}
\label{sec:inferencia}

{This section presents the inference stage, in which YOLOv8s is applied to UAV images not used for training or validation to detect cattle. The feed bunk mask obtained using SAM provides a geometric reference for quantifying cattle dispersion through statistical and geometric metrics.}

Based on the comparative results discussed in the previous section, \textbf{YOLOv8s was adopted for the inference and dispersion analysis stages}. Although YOLOv11s achieved slightly higher average detection metrics, YOLOv8s showed competitive performance while offering a consolidated implementation. Therefore, YOLOv8s was considered a suitable and reliable detector for cattle-head localization in the proposed approach.

\subsection{Analysis of cattle dispersion relative to the feed bunk}
\label{sec:feedbunk}

In addition to identifying the cattle, another required task is the identification of the feed bunk in order to infer a line equation that represents its spatial location. Due to scope limitations, a pre-trained image segmentation model known as SAM was employed.

After running the model on these images, the segmentation masks are obtained. These masks provide a polygon whose boundary represents a reasonably accurate approximation of the spatial location of the feed bunk. From the extremities of this polygon, it is possible to define a rectangle whose vertical axis allows the extraction of the line equation representing the feed bunk.

To assess the consistency of the feed bunk segmented by SAM, the fitted line obtained from the segmented mask was compared with manual annotations. The resulting angular deviation was relatively small ($0.44 \pm 0.68^\circ$), indicating that SAM provides a reasonably consistent reference for the proposed spatial analysis under the evaluated conditions.

The spatial analysis was first illustrated with two representative
images, referred to as the {regular and irregular spatial distribution scenarios, and then extended to a set of 18 UAV images with at least five detected heads, not used for detector training or validation, covering multiple acquisition campaigns and different occupation patterns of the feed bunk.
However, some of these images were acquired on the same days as training or validation images and may depict similar scenes, although these specific frames were not annotated.
Here, regularity refers to relative differences in cattle alignment and spacing, without implying feeding quality.}
For each image, cattle heads were detected with the selected YOLOv8 \textit{small} model, the feed bunk was segmented with SAM, and the transversal and longitudinal metrics defined in Section~\ref{sec:inferencias_dispersao} were computed from the central points of the detected bounding boxes. The images were then grouped according to the DBSCAN clustering of the projected head positions: {6 images presented no cluster under the adopted settings and 12 presented at least one cluster (localized aggregation along the bunk).} This grouping provides an image-level indicator of feed bunk occupation and allows the metrics to be summarized beyond the {two illustrative} examples. For these images, inference was performed using the trained model to identify the central points corresponding to the positions of each animal’s head.

Combined with the obtained line equation, the positions of the detected objects can be mapped onto a Cartesian plane in which the origin is located at the upper-left corner of the image and each unit corresponds to one pixel of the original image.
Once this mapping is established, it becomes possible to apply the equations introduced in Section \ref{sec:inferencias_dispersao}.

Finally, with the inferred points and the calculated feed bunk line equation, Equations \ref{eq:distancia-ponto-reta} and \ref{eq:proj-ortogonal} are applied to quantify instantaneous cattle spatial distribution relative to the feed bunk. Equation \ref{eq:distancia-ponto-reta} computes the orthogonal distance of each animal to the feed bunk, indicating individual spatial displacement from the feed bunk, while Equation \ref{eq:proj-ortogonal} projects the points onto the feed bunk line, enabling the analysis of longitudinal distribution along its length. Additionally, the longitudinal coordinates $s_i$ representing the projected points in one dimension are computed using Equation \ref{eq:conjunto-s}, allowing the evaluation of the mean and variability of the gaps between animals, as well as clustering analysis using DBSCAN.

\begin{figure*}[!ht]
\centering
\resizebox{.9\textwidth}{!}{\begin{minipage}{\textwidth}
\centering
\begin{subfigure}{0.47\linewidth}
\centering
\includegraphics[width=\linewidth]{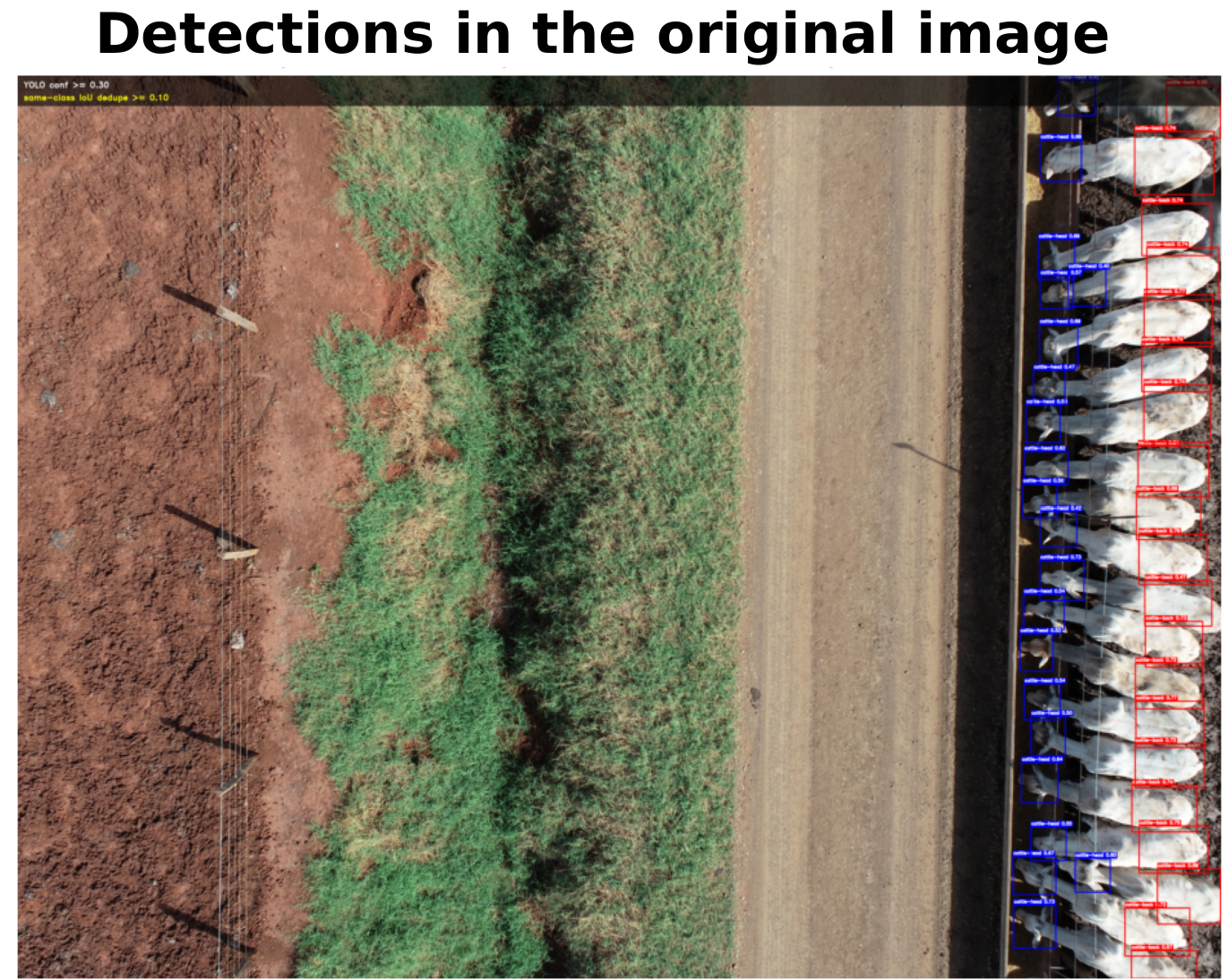}
\end{subfigure}
\hfill
\begin{subfigure}{0.47\linewidth}
\centering
\includegraphics[width=\linewidth]{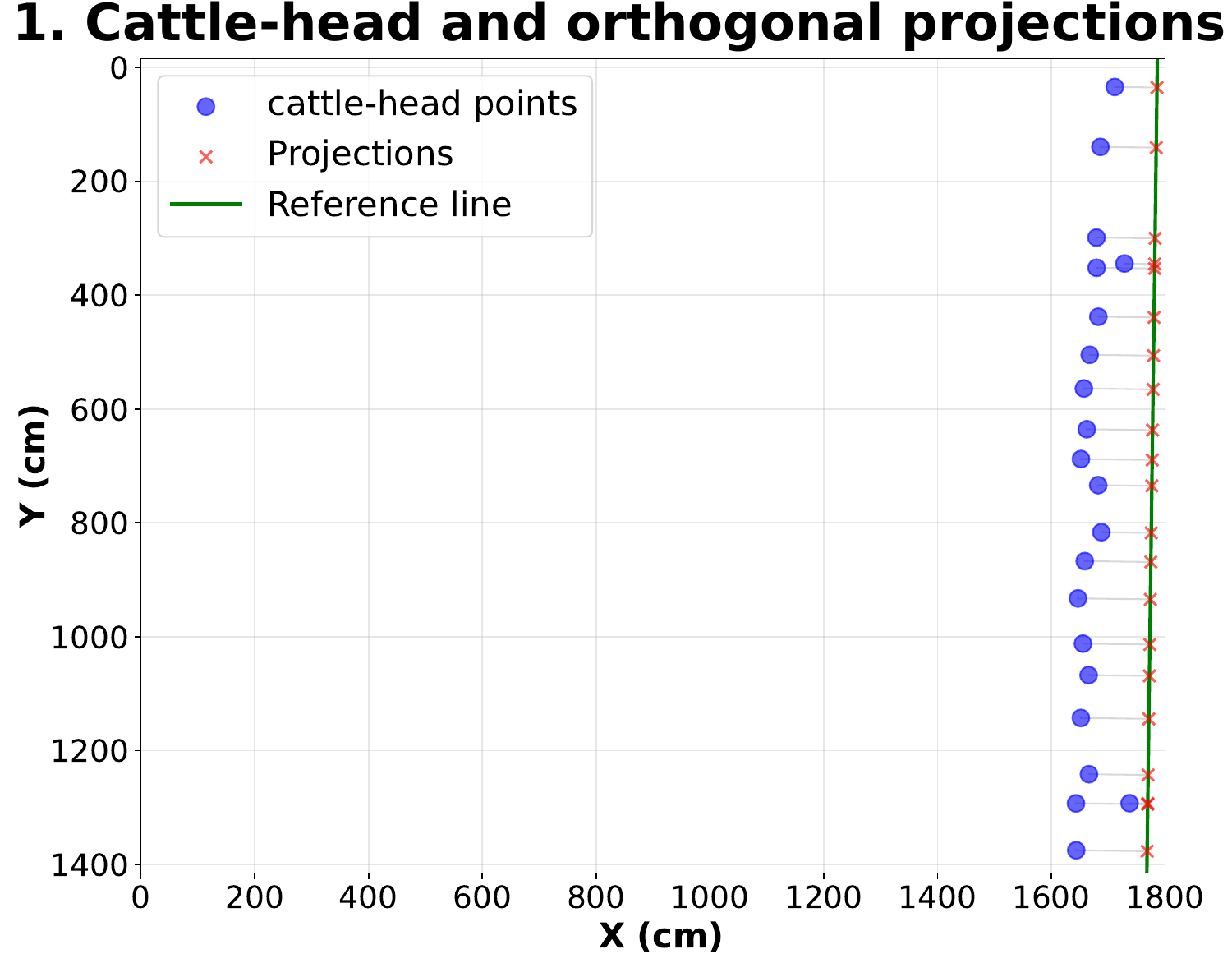}
\end{subfigure}
\vspace{1em}

\begin{subfigure}{0.49\linewidth}
\centering
\includegraphics[width=\linewidth]{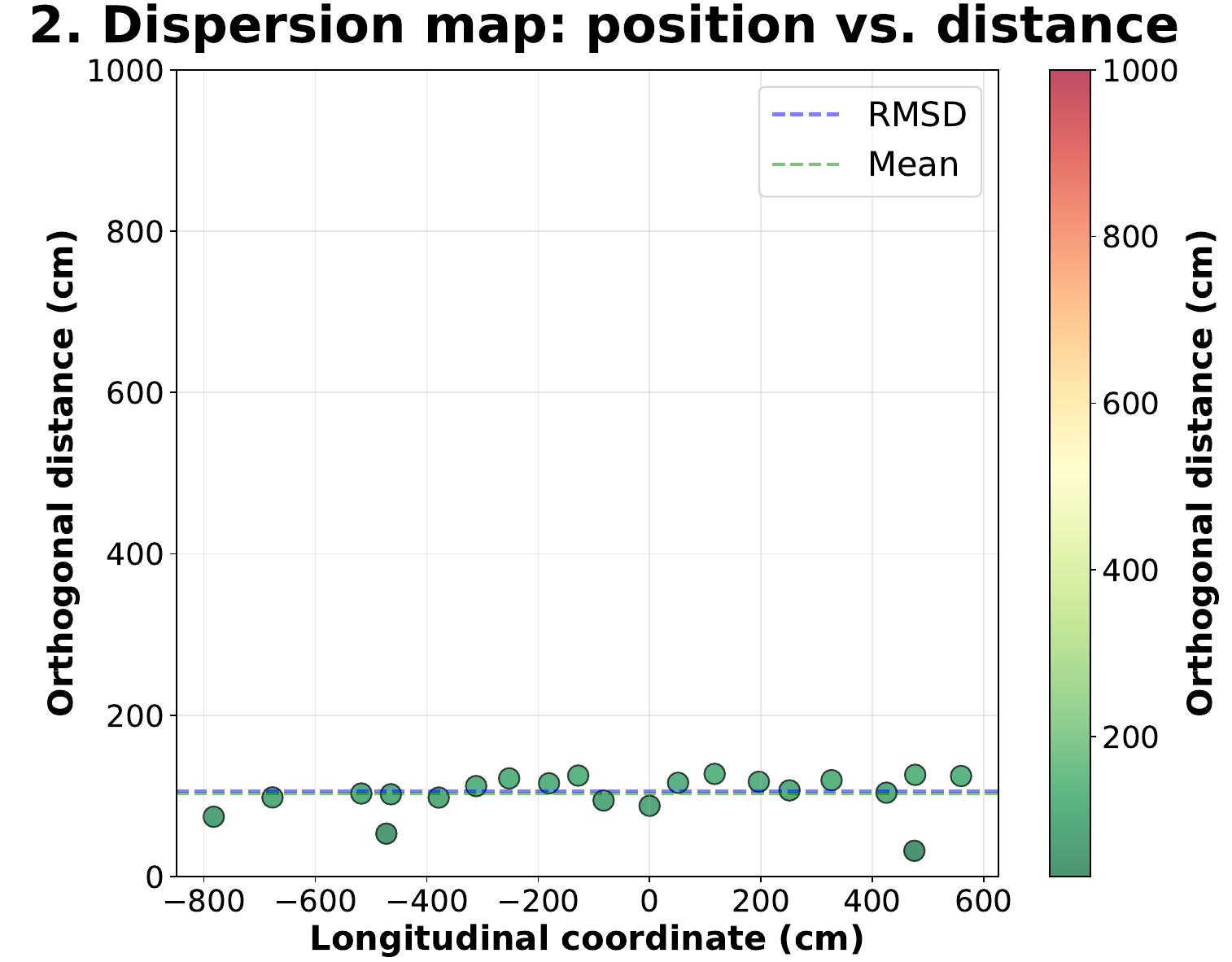}
\end{subfigure}
\hfill
\begin{subfigure}{0.47\linewidth}
\centering
\includegraphics[width=\linewidth]{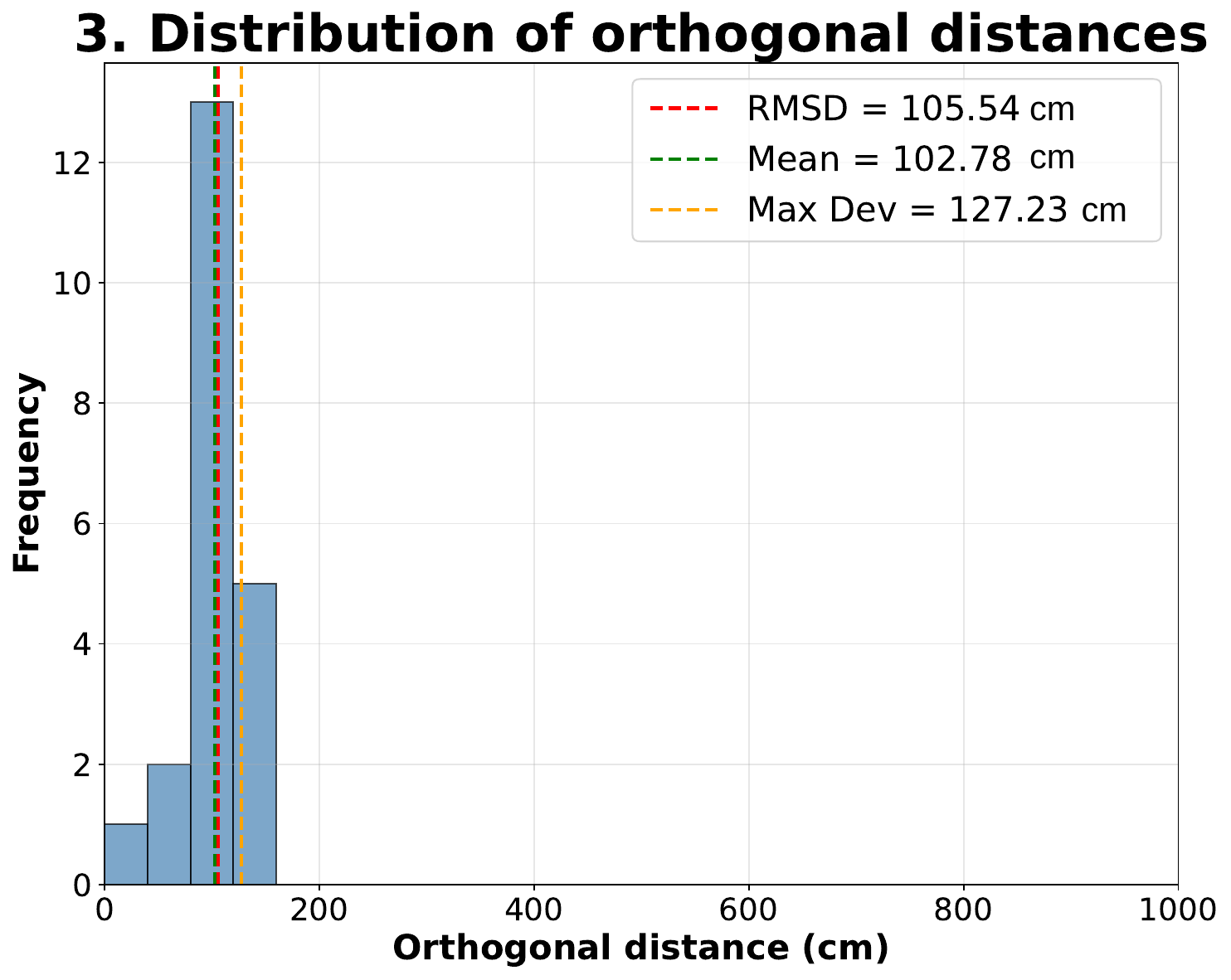}
\end{subfigure}
\vspace{1em}

\begin{subfigure}{0.44\linewidth}
\centering
\includegraphics[width=\linewidth]{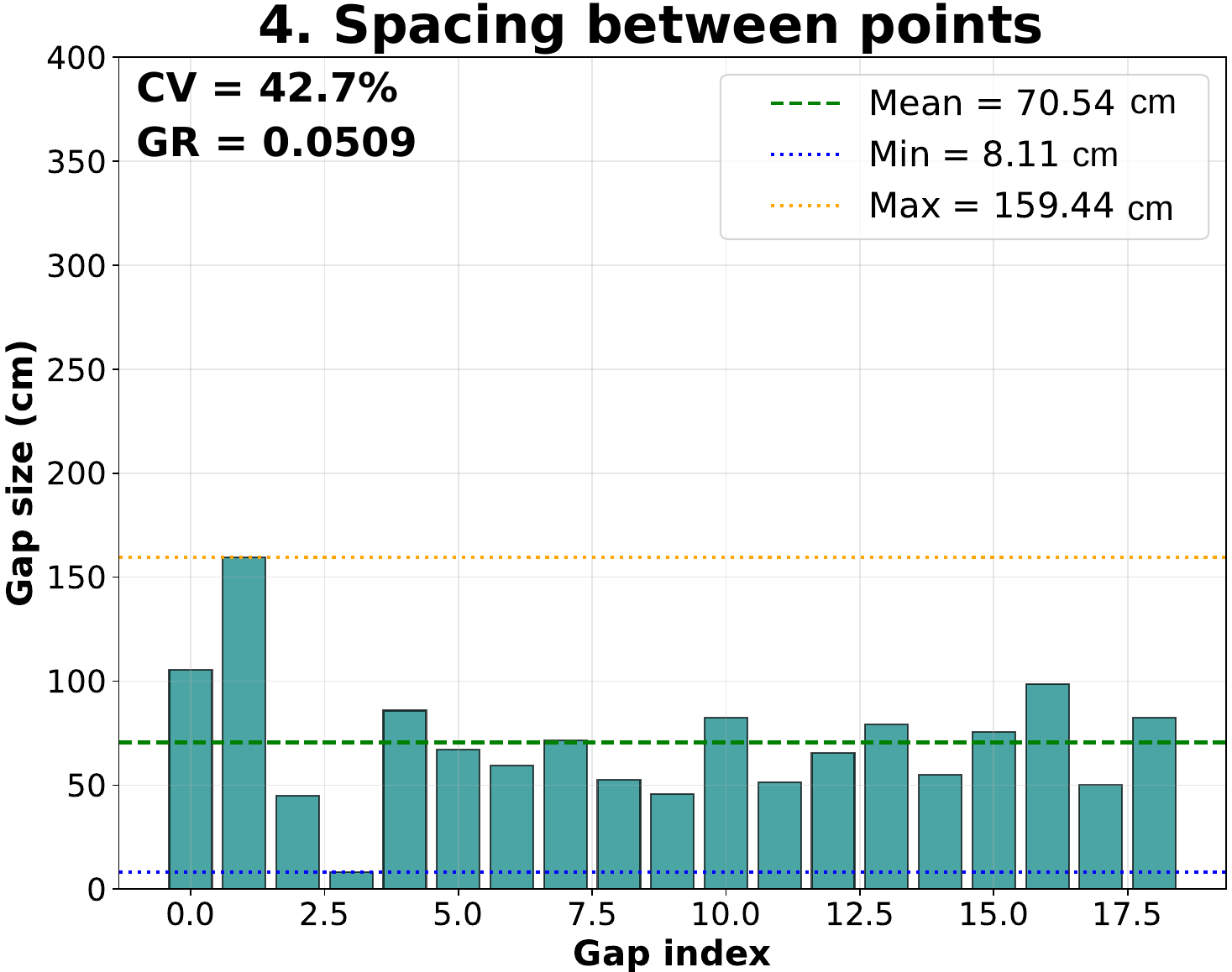}
\end{subfigure}
\hfill
\begin{subfigure}{0.47\linewidth}
\centering
\includegraphics[width=\linewidth]{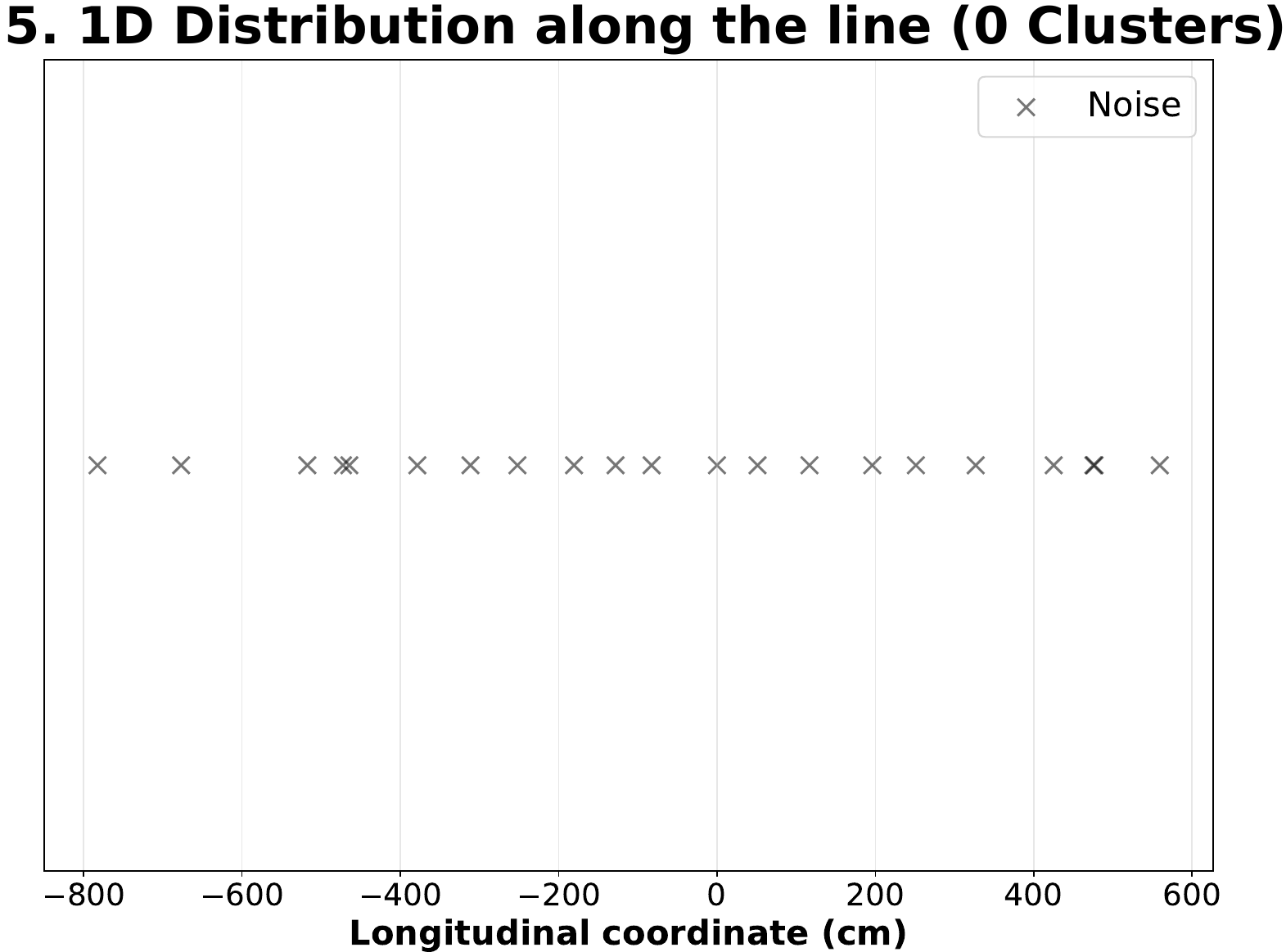}
\end{subfigure}
\end{minipage}}
\caption{{Dashboard summarizing the spatial distribution metrics for an example with regular spatial distribution, characterized by more even longitudinal coverage and consistent proximity to the feed bunk. The relatively low RMSD (105.54 cm), limited maximum deviation (127.23 cm), moderate CV (42.70\%) and absence of clusters indicate a more regular pattern.}}
\label{fig:results-bom}
\end{figure*}

\begin{figure*}[!ht]
\centering
\resizebox{.9\textwidth}{!}{\begin{minipage}{\textwidth}
\centering
\begin{subfigure}{0.47\linewidth}
\centering
\includegraphics[width=\linewidth]{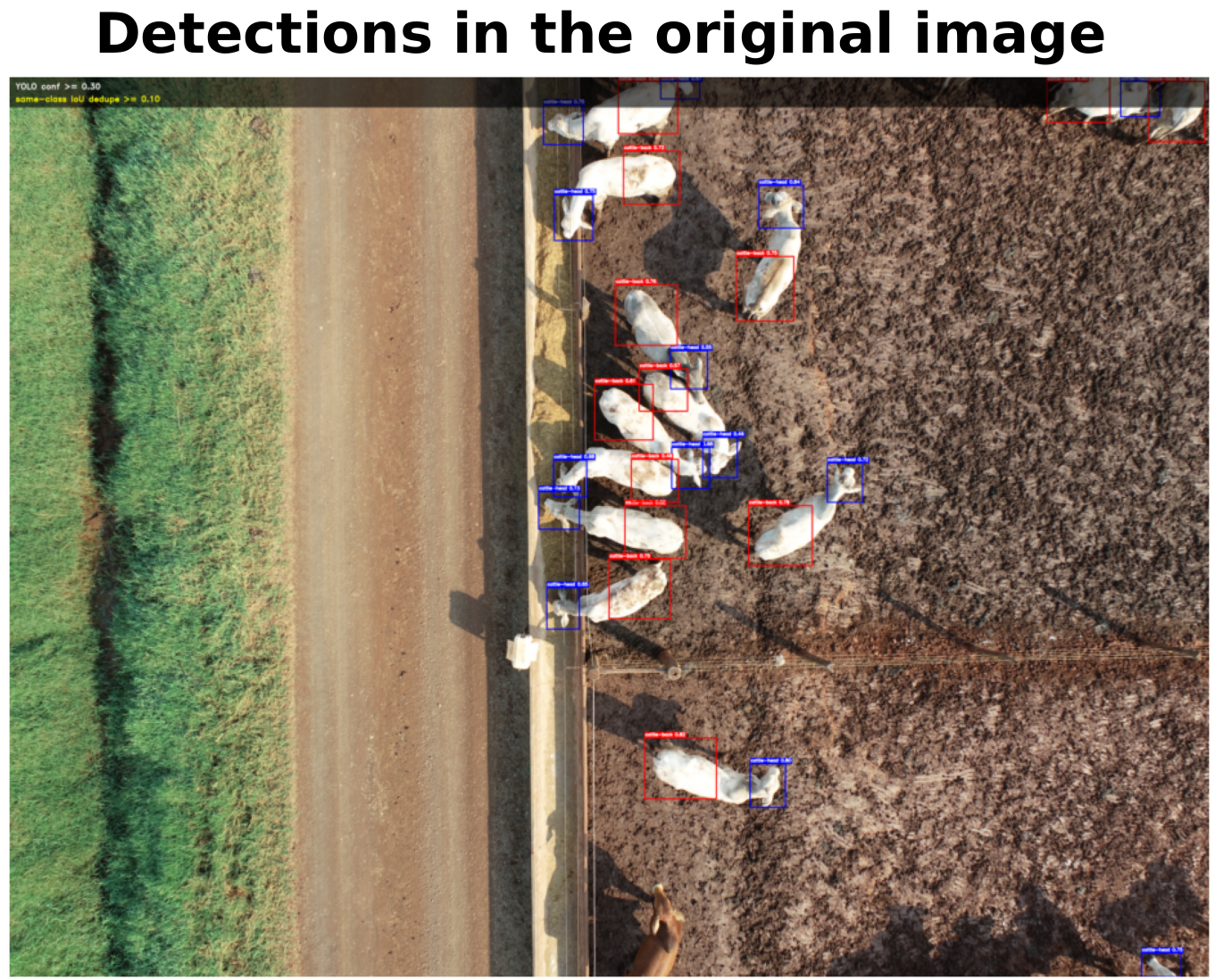}
\end{subfigure}
\hfill
\begin{subfigure}{0.48\linewidth}
\centering
\includegraphics[width=\linewidth]{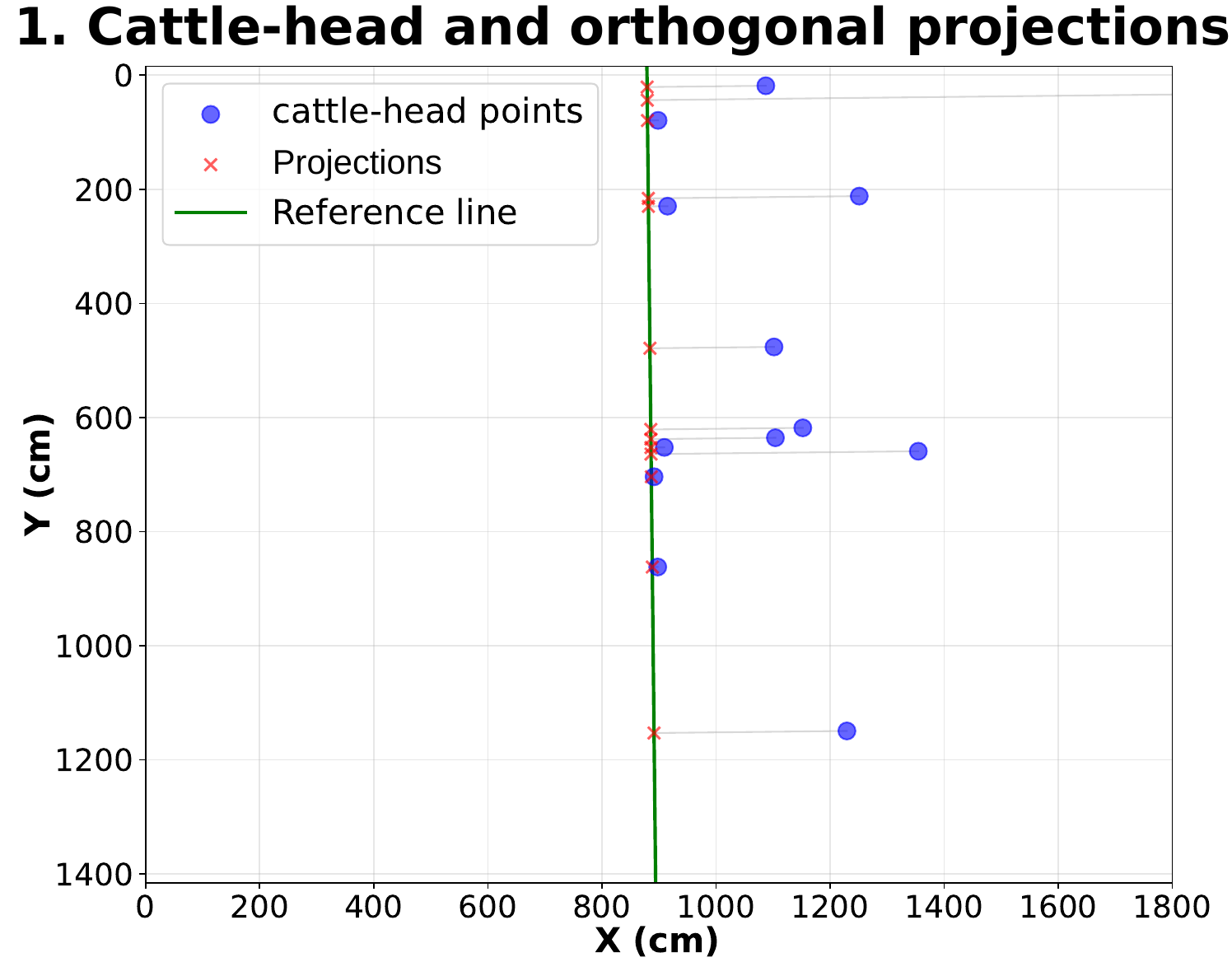}
\end{subfigure}
\vspace{1em}

\begin{subfigure}{0.49\linewidth}
\centering
\includegraphics[width=\linewidth]{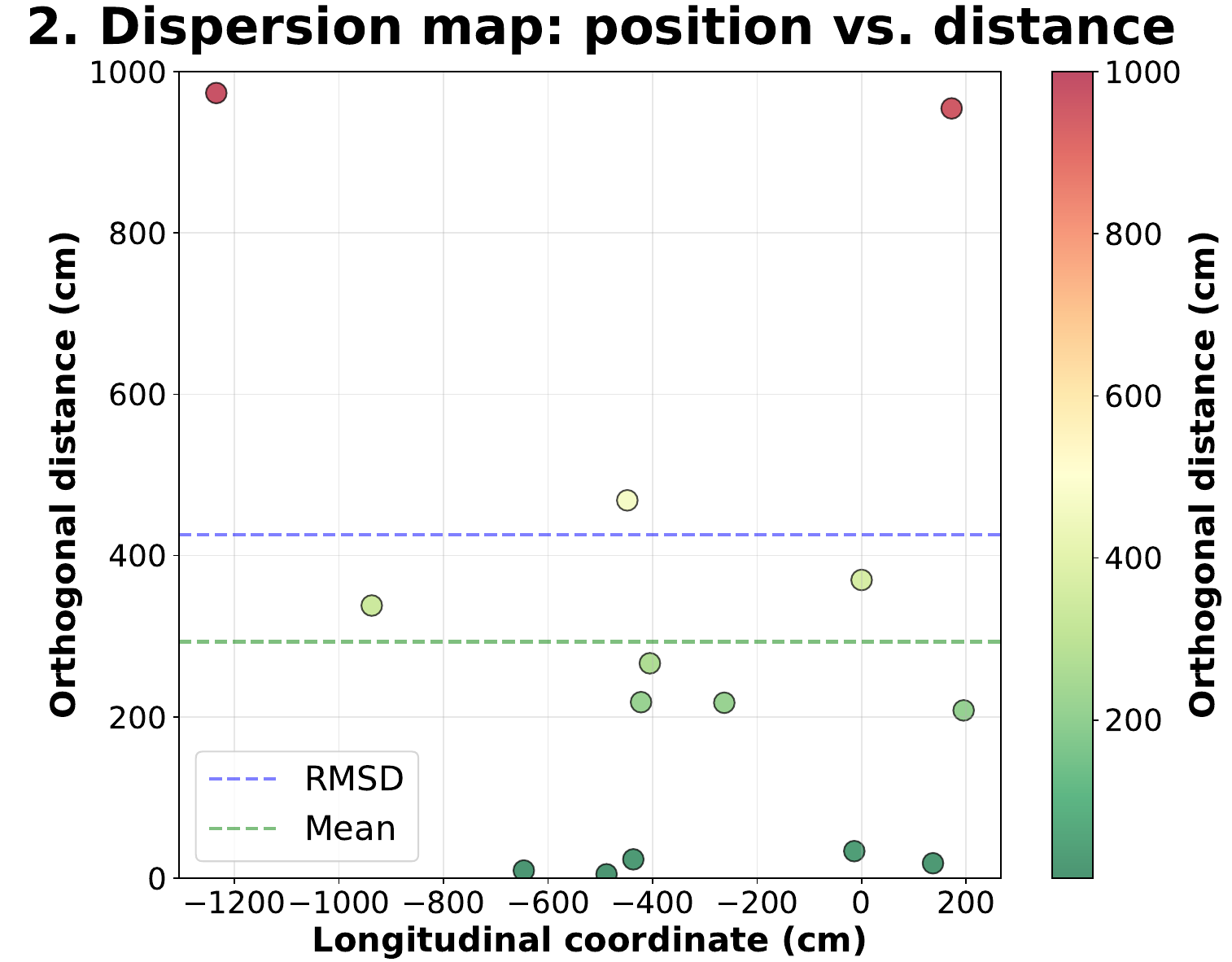}
\end{subfigure}
\hfill
\begin{subfigure}{0.48\linewidth}
\centering
\includegraphics[width=\linewidth]{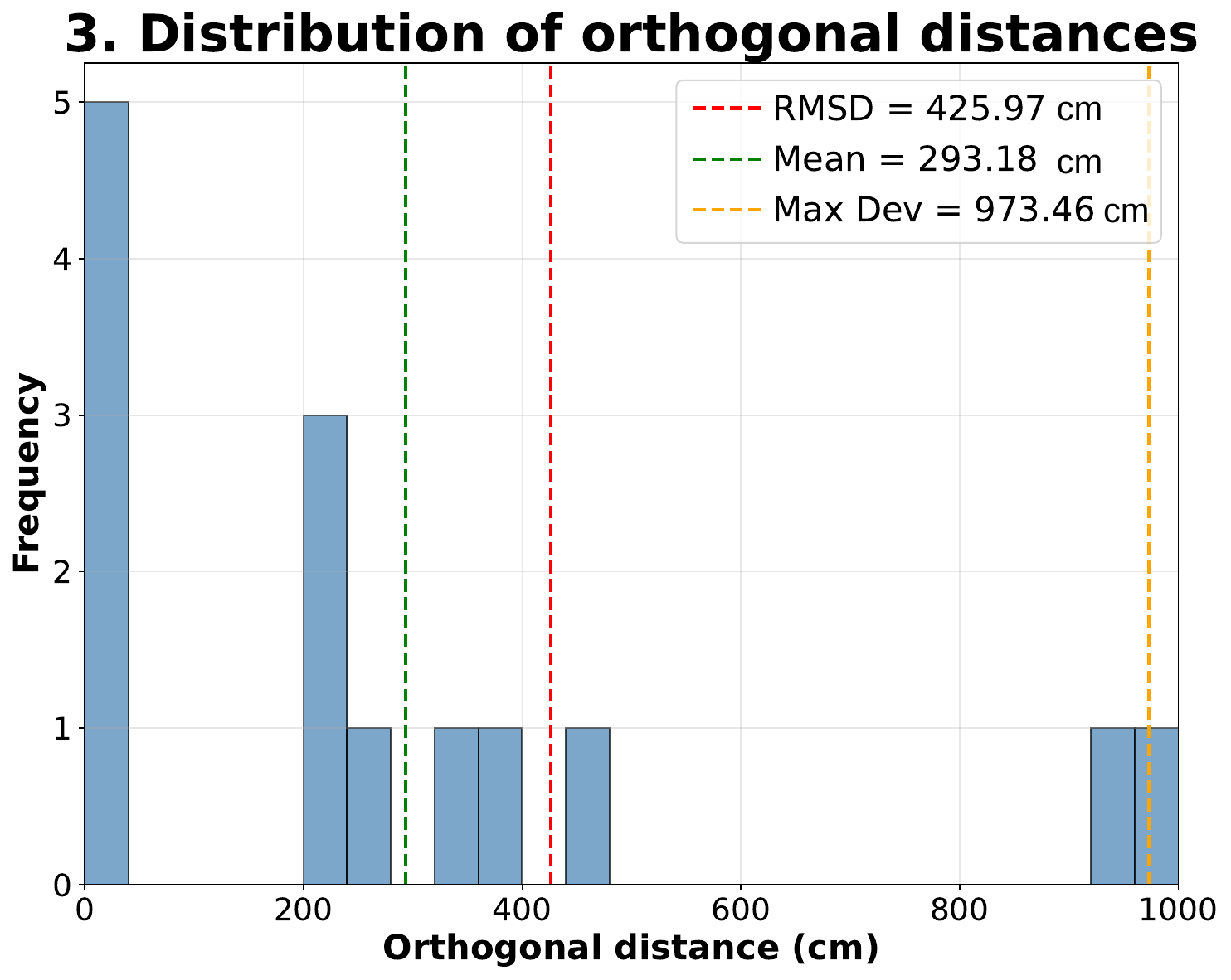}
\end{subfigure}
\vspace{1em}

\begin{subfigure}{0.44\linewidth}
\centering
\includegraphics[width=\linewidth]{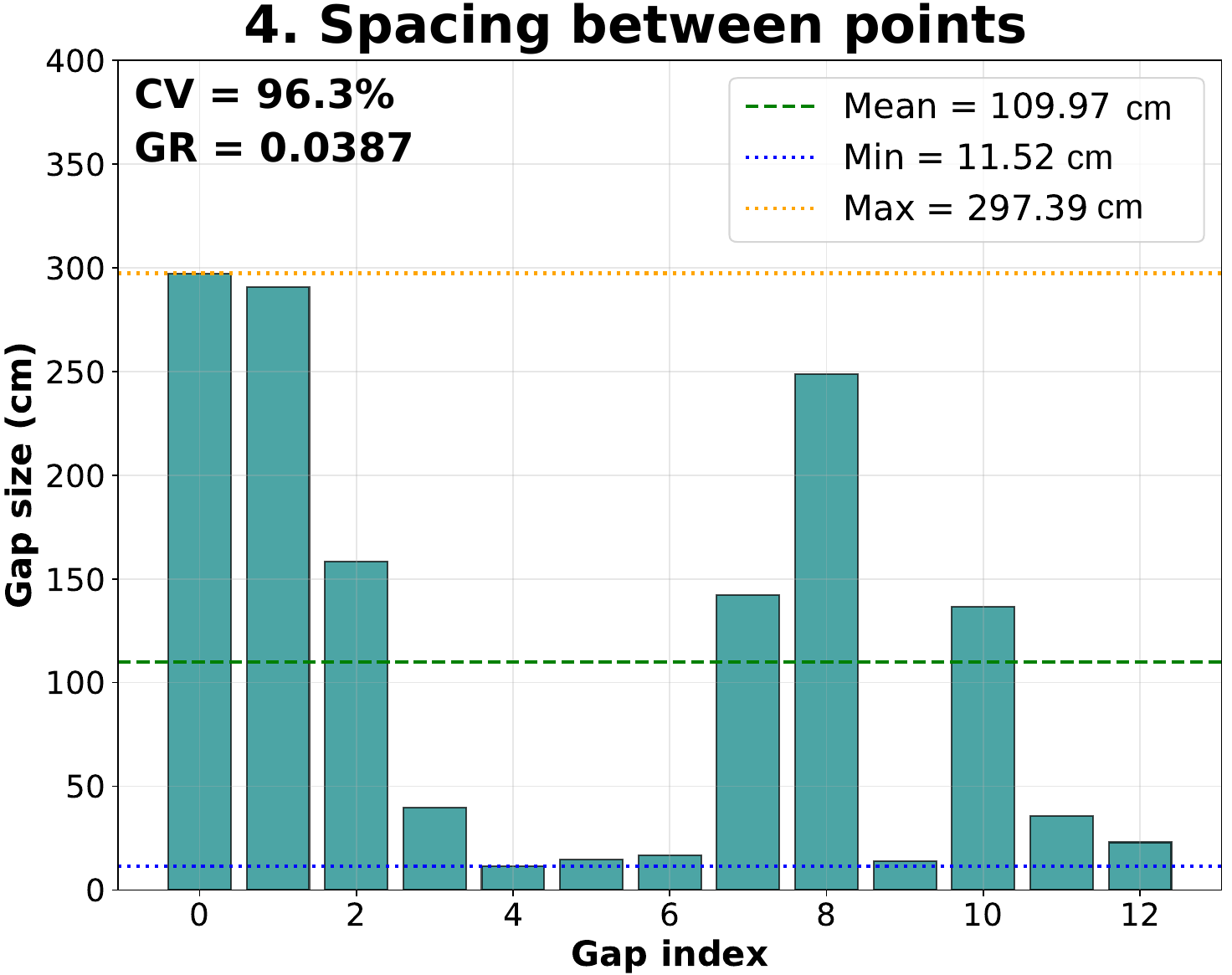}
\end{subfigure}
\hfill
\begin{subfigure}{0.47\linewidth}
\centering
\includegraphics[width=\linewidth]{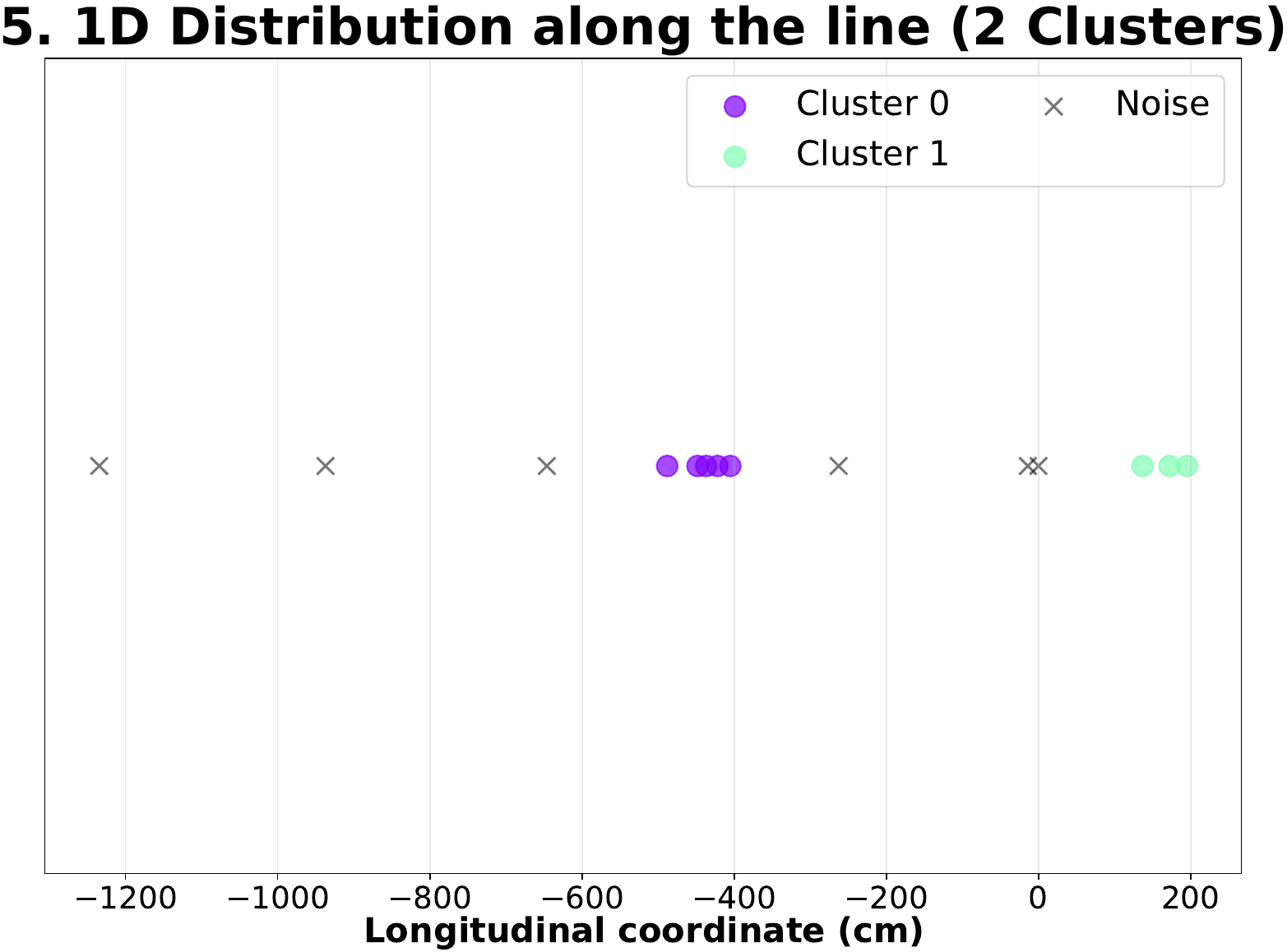}
\end{subfigure}
\end{minipage}}
\caption{{Dashboard summarizing the spatial distribution metrics for an example with irregular spatial distribution, characterized by uneven coverage and large distances from the feed bunk. The substantially higher RMSD (425.97 cm), maximum deviation (973.46 cm), elevated CV (96.31\%), low GR (0.0387), and two clusters indicate fragmented spacing and local aggregation.}}
\label{fig:results-ruim}
\end{figure*}

{Figures~\ref{fig:results-bom} and~\ref{fig:results-ruim} present dashboards that summarize these metrics for two representative scenarios, respectively illustrating regular spatial distribution (more even longitudinal coverage and consistent proximity to the feed bunk) and irregular spatial distribution (uneven coverage, large deviations, and clustering).
The regular scenario is characterized by cattle positioned closer to the feed bunk and more evenly spaced along its extension. This configuration is reflected by lower values in the orthogonal distance metrics and lower variability in the longitudinal spacing between adjacent animals.}

{We used a DBSCAN radius of half the mean spacing ($\epsilon=0.50\,\mu_g$) and a minimum of three points for dense neighborhoods (minPts $=3$). These settings target local concentrations of animals and prevent isolated pairs from forming clusters. Larger $\epsilon$ or smaller minPts can increase the proportion of images with clusters, affecting the resulting image grouping.}

Tables~\ref{tabela:metricas-dist} and~\ref{tabela:metricas-gaps} report the quantitative values corresponding to the dashboards shown in Figures~\ref{fig:results-bom} and~\ref{fig:results-ruim}. In Table \ref{tabela:metricas-dist}, the mean orthogonal distance indicates the typical separation of animals from the feed bunk, while the RMSD penalizes larger deviations, being more sensitive to animals that are significantly distant. The maximum deviation identifies the most distant animal, revealing extreme cases of spatial displacement from the feed bunk. Low values for these metrics indicate close spatial alignment with the feed bunk.

\begin{table}[!ht]
\centering
\caption{{Transversal dispersion metrics: orthogonal distances of animals relative to the feed bunk. The regular and irregular scenarios illustrate spatial distributions (single-image values), while the without/with cluster rows report the mean over feed bunk images. All values are in centimeters (GSD $\approx 4$~mm/px).}}
\label{tabela:metricas-dist}
\resizebox{\columnwidth}{!}{
{
\begin{tabular}{l|c|c|c|c}
\hline
\multicolumn{1}{c|}{\textbf{\shortstack{Image /}}} & \textbf{Mean} & \textbf{Standard} & \textbf{RMSD} & \textbf{Maximum} \\
\multicolumn{1}{c|}{\textbf{Group}} & \textbf{} & \textbf{Deviation} & \textbf{} & \textbf{Deviation} \\ \hline \hline
Regular scenario & 102.78 & 23.97  & 105.54 & 127.23  \\
Irregular scenario & 293.18 & 309.03 & 425.97 & 973.46  \\ \hline
Without cluster  & 146.66 & 59.31  & 163.41 & 262.43  \\
With cluster     & 470.72 & 393.68 & 615.30 & 1,104.04 \\ \hline
\end{tabular}
}
}
\end{table}

Table \ref{tabela:metricas-gaps} evaluates longitudinal distribution uniformity through the spacing (\textit{gaps}) between consecutive animals projected onto the feed bunk line. The Coefficient of Variation (CV) quantifies spacing regularity: {low values indicate a uniform distribution, while high values indicate irregular spacing}. The Gap Ratio (GR) complements this analysis by computing the ratio between the smallest and largest spacing, where values close to 1.0 indicate similar gap sizes and values close to 0 indicate strong spacing imbalance. The number of clusters detected by the DBSCAN algorithm identifies {local aggregations: zero clusters indicate no aggregation detected under the adopted settings, while one or more clusters reveal concentrated group formation.} 

\begin{table}[!ht]
\centering
\caption{{Longitudinal uniformity metrics describing the spacing between animals along the feed bunk. The regular and irregular scenarios illustrate spatial distributions (single-image values), while the without/with cluster rows report means, except for cluster counts, which are summarized by the mode. Mean and standard deviation in centimeters (GSD $\approx 4$~mm/px) and the remaining metrics are dimensionless.}}
\label{tabela:metricas-gaps}
\resizebox{\columnwidth}{!}{
{
\begin{tabular}{l|c|c|c|c|c}
\hline
\multicolumn{1}{c|}{\textbf{Image /}} & \textbf{Mean} & \textbf{Standard} & \textbf{Clusters} & \textbf{CV} & \textbf{GR} \\
\multicolumn{1}{c|}{\textbf{Group}} & \textbf{} & \textbf{Deviation} & \textbf{(mode)} & \textbf{} & \textbf{} \\ \hline \hline
Regular scenario & 70.54  & 30.12  & 0    & 42.70\% & 0.0509 \\
Irregular scenario & 109.97 & 105.92 & 2    & 96.31\% & 0.0387 \\ \hline
Without cluster  & 152.76 & 72.22  & 0    & 52.16\% & 0.1690 \\
With cluster     & 88.28  & 75.44  & 1  & 86.05\% & 0.0395 \\ \hline
\end{tabular}
}
}
\end{table}

The grouped results show that images with clusters presented higher transversal dispersion than images without clusters, with RMSD increasing {from $163.41 \pm 94.05$~cm to $615.30 \pm 223.39$~cm. Longitudinal irregularity also increased in clustered images, with CV rising from $52.16 \pm 24.97\%$ to $86.05~\pm~21.57\%$, indicating less uniform spacing along the feed bunk.}
{Given the limited sample and variability across images, these results should be interpreted as feasibility trends rather than findings from statistically powered comparisons.}

{Comparing the presented scenarios, the image with regular spatial distribution exhibits an RMSD of $105.54$~cm and a maximum deviation of $127.23$~cm, in contrast to $425.97$~cm and $973.46$~cm in the irregular distribution image, approximately $4.04$ and $7.65$ times larger, respectively.
This disparity indicates that, in the irregular scenario, animals are not only farther away on average, but some individuals are positioned extremely far from the feed bunk.
Since the distance metrics scale linearly with GSD, a hypothetical $\pm~10\%$ variation in this factor would change the RMSD by $\pm~10.55$~cm and $\pm~42.60$~cm in the regular and irregular scenarios, respectively.
Regarding longitudinal distribution, the regular image presents a lower CV of 42.70\% and no cluster formation, although its low GR (0.0509) indicates local gap imbalance, while the irregular image shows a CV of 96.31\%, a GR of only 0.0387, and the formation of two distinct clusters.}
{Notice that GR tends to saturate near zero in clustered cases, making it less sensitive to clustering severity and somewhat redundant with CV in indicating spacing imbalance. Therefore, it is retained only for completeness.}

These metrics are visualized through the dashboards generated for each image (Figures \ref{fig:results-bom} and \ref{fig:results-ruim}), which present complementary visualizations: (1) original points and their orthogonal projections onto the feed bunk line, (2) a dispersion map relating longitudinal position and orthogonal distance to the feed bunk, (3) a histogram of orthogonal distances with markers for RMSD and maximum deviation, (4) analysis of spacing between consecutive animals, and (5) a one-dimensional distribution along the feed bunk with cluster identification.
These results objectively quantify the qualitative difference between two opposing feed bunk spatial distribution scenarios and can support the analysis of instantaneous cattle alignment and spacing at the feed bunk.

\section{Conclusion}
\label{sec:conclusion}

This paper investigated UAV-based computer vision to support Precision Livestock Farming, focusing on cattle detection and assessing feed bunk alignment and instantaneous spatial distribution under real-world feedlot conditions. A UAV dataset annotated into \textit{cattle-head} and \textit{cattle-back} classes enabled not only effective cattle detection under the evaluated conditions but also a more interpretable analysis of animal spatial arrangement relative to the feed bunk.

{By modeling the segmented feed bunk as a geometric reference and computing orthogonal distances and longitudinal projections of detected cattle, the proposed framework translates qualitative spatial distribution patterns observed during feeding into objective and interpretable metrics, including dispersion, spacing regularity, and clustering.}
Experimental results on contrasting scenarios and UAV feed bunk images indicated that these metrics are sensitive to differences in alignment and spatial dispersion, discriminating more regular occupation patterns from irregular or clustered ones under the evaluated conditions.

A limitation of this study is the relatively small image-level dataset, so the results should be interpreted as feasibility evidence under the evaluated feedlot conditions.
{The regular and irregular spatial distribution scenarios were defined by our qualitative visual assessment, without independent validation against expert bunk scores or zootechnical measurements.}
The explicit detection of the head region also supports extensions toward future temporal analyses, such as estimating how long animals remain engaged in feeding.
{In future work, we intend to extend the framework to other feed bunk layouts and validate it against expert bunk scores or zootechnical measurements across different farms, breeds, and environmental conditions.}
We also intend to extend the approach toward fully automated segmentation, animal tracking, and the integration of feed availability information to further strengthen its applicability in Precision Livestock Farming. In addition, we intend to evaluate the impact of different UAV flight altitudes on detection performance and geometric feeding metrics.

\section*{Acknowledgments}
\noindent We thank the Fazenda Campanário for providing access to the feedlot facilities and cattle.
We also express our appreciation to the farm's team for their valuable collaboration and technical support during the field data collection activities.
\noindent This work was also supported by the São Paulo Research Foundation (FAPESP) under Grants \#2023/03870-8, \#2022/09319-9, \#2024/15430-5, \#2025/10602-5; by the Brazilian National Council for Scientific and Technological Development (CNPq) under Grants \#406417/2022-9, \#403307/2025-2, and \#306788/2023-3; and by the University of São Paulo under PRPI Ordinance No.~1032 (New Faculty Support Program).

\end{document}